\documentclass[10pt,twocolumn,letterpaper]{article}

\usepackage[pagenumbers]{cvpr} 

\usepackage{graphicx}
\usepackage{amsmath}
\usepackage{amssymb}
\usepackage{booktabs}
\usepackage{xcolor}
\usepackage{overpic}

\usepackage{bm}
\usepackage{enumitem}
\setlist[enumerate]{itemsep=0mm}
\usepackage{times}
\usepackage{mathptmx}
\usepackage{mathtools}

\usepackage{multirow}
\usepackage{makecell}

\DeclareMathAlphabet{\altmathcal}{OMS}{cmsy}{m}{n}
\DeclareMathAlphabet{\mathbfit}{OT1}{ptm}{bx}{it}

\newlength\paramargin
\newlength\figmargin

\newlength\secmargin
\newlength\figcapmargin
\newlength\tabcapmargin

\usepackage{soul}
\usepackage{microtype}

\usepackage{chngcntr}
\usepackage[scaled]{beramono}
\usepackage[T1]{fontenc}

\usepackage[pagebackref,breaklinks,colorlinks]{hyperref}

\usepackage[capitalize]{cleveref}
\crefname{section}{Sec.}{Secs.}
\Crefname{section}{Section}{Sections}
\Crefname{table}{Table}{Tables}
\crefname{table}{Tab.}{Tabs.}

\def\cvprPaperID{815} 
\def\confName{CVPR}
\def\confYear{2022}

\newcommand{\xpos}{\mathbf{x}}

\newcommand{\direction}{\boldsymbol{\omega}}

\newcommand{\dirout}{\vec{\direction}}

\newcommand{\tnear}{{t_\text{n}}}
\newcommand{\tfar}{{t_\text{f}}}
\newcommand{\network}{F_{\boldsymbol\theta}}

\newcommand{\firstplane}{\pi^{xy}}
\newcommand{\secondplane}{\pi^{uv}}
\newcommand{\posenc}{\boldsymbol\gamma}
\newcommand{\ray}{\mathbf{r}}
\newcommand{\rayparam}{x, y, u, v}
\newcommand{\voxset}{\altmathcal{V}}
\newcommand{\vox}{\upsilon}
\newcommand{\col}{\mathbf{c}}

\makeatletter
\renewcommand{\paragraph}{%
  \@startsection{paragraph}{4}%
  {\z@}{0.5ex \@plus .2ex \@minus .2ex}{-1em}%
  {\normalfont\normalsize\bfseries\maybe@addperiod}%
}
\newcommand{\maybe@addperiod}[1]{%
  #1\@addpunct{.}%
}
\makeatother

\newcommand{\topic}[1]{\paragraph{#1}}

\newbox\jsavebox%

\newcommand{\mpage}[2]
{
\begin{minipage}{#1\linewidth}\centering
#2
\end{minipage}
}

\newcommand{\tb}[1]{\textbf{#1}}

\begin{document}

\title{Learning Neural Light Fields with Ray-Space Embedding}

\author{%
\parbox{.2\linewidth}{\centering%
Benjamin Attal$^*$\\
{\small Carnegie Mellon University}%
}
\hspace{-3mm}
\and
\hspace{-3mm}
\parbox{.15\linewidth}{\centering%
Jia-Bin Huang\\
{\small Meta}%
}
\hspace{-3mm}
\and
\hspace{-3mm}
\parbox{.2\linewidth}{\centering%
Michael Zollh\"ofer\\
{\small Reality Labs Research}%
}
\hspace{-3mm}
\and
\hspace{-3mm}
\parbox{.15\linewidth}{\centering%
Johannes Kopf\\
{\small Meta}%
}
\hspace{-3mm}
\and
\hspace{-3mm}
\parbox{.15\linewidth}{\centering%
Changil Kim\\
{\small Meta}%
}
}

\twocolumn[{
\renewcommand\twocolumn[1][]{#1}
\maketitle
\vspace{-10mm}
\begin{center}
\textbf{\url{https://neural-light-fields.github.io}}
\end{center}
\vspace{2mm}
\begin{center}
\vspace{-6mm}
\includegraphics[width=0.71\linewidth]{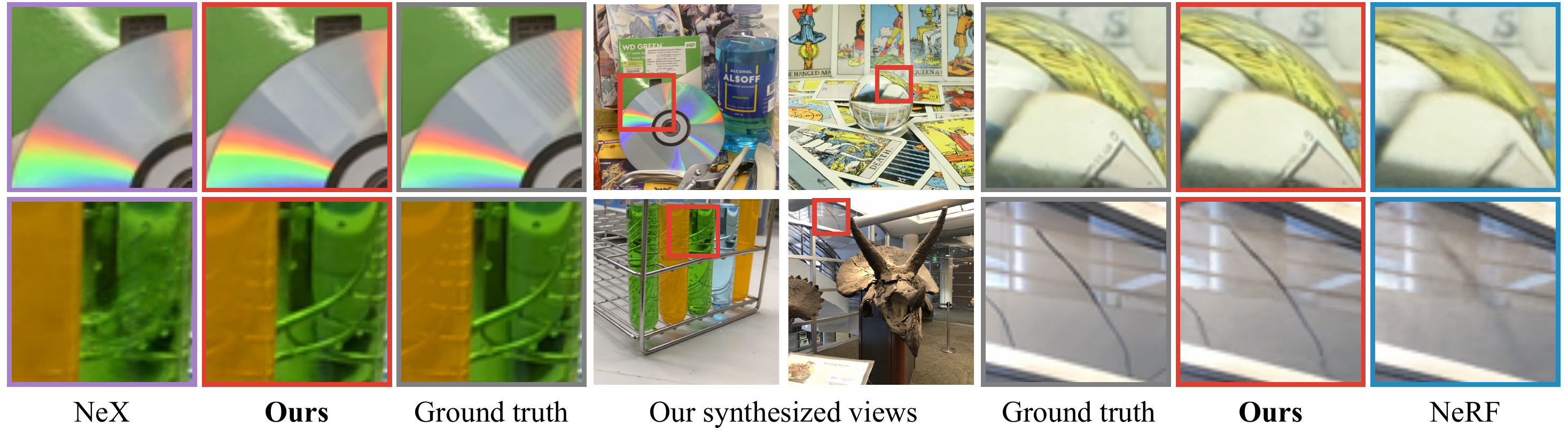}\hfill%
\includegraphics[width=0.285\linewidth]{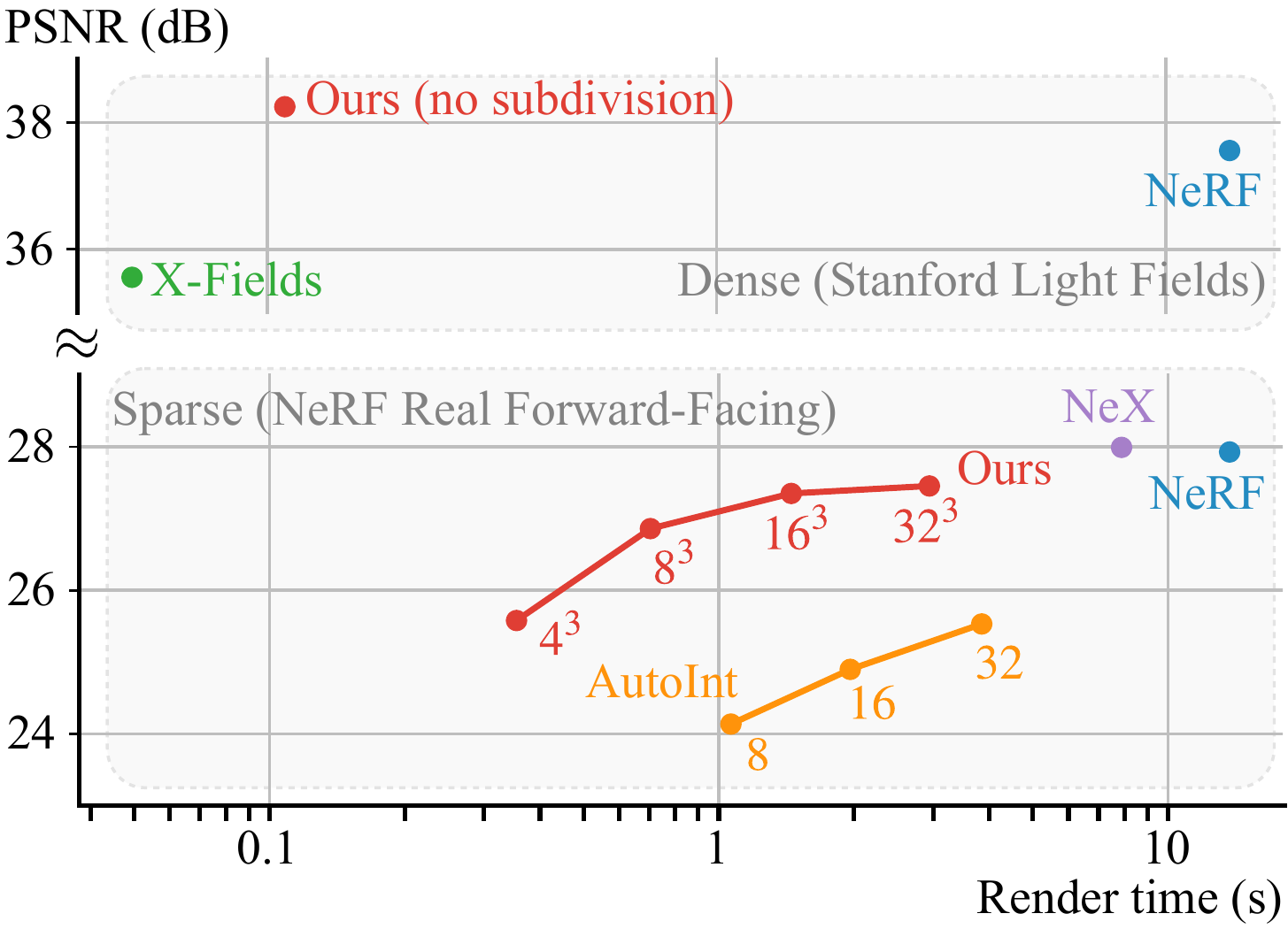}
%
%
%
%
%

%
\vspace{-1mm}
\captionof{figure}{%
We present \emph{Neural Light Fields with Ray-Space Embedding}, which map rays directly to integrated radiance. Our representation allows for high-quality view synthesis with faithful reconstruction of complex view dependence (\emph{left}). We are able to synthesize novel views in a fraction of the time required for the current state-of-the-art approaches, with comparable memory footprint.  Our method achieves a more favorable trade-off between quality, speed, and memory than existing approaches for both dense and sparse light fields (\emph{right}).
See more visual comparisons including video results in our \emph{project website} linked above.%
}
\vspace{2mm}
\label{fig:teaser}
\end{center}

}]

\maketitle

\thispagestyle{empty}
\begin{abstract}

\noindent
Neural radiance fields (NeRFs) produce state-of-the-art view synthesis results, but are slow to render, requiring hundreds of network evaluations per pixel to approximate a volume rendering integral.
Baking NeRFs into explicit data structures enables efficient rendering, but results in large memory footprints and, in some cases, quality reduction.
Additionally, volumetric representations for view synthesis often struggle to represent challenging view dependent effects such as distorted reflections and refractions.
We present  a novel neural light field representation that, in contrast to prior work, is fast, memory efficient, and excels at modeling complicated view dependence.
Our method supports rendering with a single network evaluation per pixel for small baseline light fields and with only a few evaluations per pixel for light fields with larger baselines.
At the core of our approach is a ray-space embedding network that maps 4D ray-space into an intermediate, interpolable latent space.
Our method achieves state-of-the-art quality on dense forward-facing datasets such as the Stanford Light Field dataset. 
In addition, for forward-facing scenes with sparser inputs we achieve results that are competitive with NeRF-based approaches while providing a better speed/quality/memory trade-off with far fewer network evaluations.
{\let\thefootnote\relax\footnote{{\hspace{-5mm}$^*$ This work was done while Benjamin was an intern at Meta.}}}

\end{abstract}

\section{Introduction}
\label{sec:intro}

\noindent
View synthesis is an important problem in computer vision and graphics.
Its goal is to photorealistically render a scene from unobserved camera poses, given a few posed input images.
Existing approaches solve this problem by optimizing some underlying representation of the scene's appearance and geometry and then rendering this representation from novel views.

View synthesis has recently experienced a renaissance with an explosion of interest in \textit{neural scene representations} --- see Tewari et al.~\cite{tewari2020state,tewari2021NeuralAdvances} for a snapshot of the field.
\textit{Neural radiance fields}~\cite{mildenhall2020nerf} are perhaps the most popular of these neural representations, and methods utilizing them have recently set the state-of-the-art in rendering quality for view synthesis.
A radiance field is a 5D function that maps a 3D point $\xpos$ and 3D direction $\dirout$ (with only 2 degrees of freedom) to the radiance leaving $\xpos$ in direction $\dirout$, as well as the density of the volume at point $\xpos$. A neural radiance field or NeRF represents this function with a neural network.
Because volume rendering a NeRF is differentiable, it is straightforward to optimize by minimizing the difference between ground truth views at known camera poses and their reconstructions.
%

The main drawback of neural radiance fields is that volume rendering requires many samples and thus many neural network evaluations per ray to approximate a volume rendering accurately.
Thus, rendering from a NeRF is usually quite slow.
Various approaches exist for baking or caching neural radiance fields into explicit data structures to improve efficiency~\cite{yu2021plenoctrees,hedman2021baking,reiser2021kilonerf,Garbin21arxiv_FastNeRF}.
Some approaches, concurrent to this work, learn color and density directly on a voxel grid, which improves both training and rendering speed~\cite{yu2021plenoxels,muller2022instant,sun2021direct}.
However, the storage cost for explicit representations is much higher than a NeRF. Further, the baking procedure itself sometimes leads to a loss in resulting view synthesis quality for baking methods.
Other methods aim to reduce the number of neural network evaluations per ray by representing radiance only on surfaces~\cite{Kellnhofer2021nlr,nef2021donerf}.
These methods predict new images with only a few evaluations per ray. However, their quality is contingent on either ground truth geometry or reasonable geometry estimates, which are not always available.

A \textit{light field}~\cite{levoy1996light,gortler1996lumigraph} is the integral of a radiance field.
It maps ray parameters directly to the integrated radiance along that ray. Thus, only one look-up of the underlying representation is required to determine the color of a ray; hence one evaluation per pixel, unlike hundreds of evaluations required by a NeRF.
A common assumption for light fields is that this integral remains the same no matter the ray origin (i.e., radiance is constant along rays), which holds when the convex hull of the scene geometry does not contain any viewpoints used for rendering \cite{levoy1996light}.
Given this assumption, a light field is a function of a ray on a 4D ray space.
%

In this paper, we show how to learn \emph{neural light fields}. 
%
%
Since coordinate-based neural representations have been successfully employed to learn radiance fields from a set of ground truth images, one might think that they could be useful for representing and learning light fields as well.
However, we show that learning light fields is significantly more challenging than learning radiance fields. 
Using the same neural network architecture as in NeRF to parameterize a light field leads to poor interpolation quality for view synthesis.

While a radiance field is a 5D function, an essential ingredient of NeRF is that it learns the scene geometry as a density field in 3D space. 
Additionally, a NeRF's learned appearance is closer to a 3D function than a 5D function since the network is late-conditioned on viewing directions \cite{zhang2020nerfpp}. 
This makes NeRFs easy to optimize but also means that they can struggle to represent complex view-dependent effects such as reflections and refractions \cite{wizadwongsa2021nex} which violate multi-view color constraints.

On the other hand, we face the problem of learning a function defined on a 4D ray space from only partial observations --- input training images only cover a few 2D slices of the full 4D space. At the same time, NeRF enjoys multiple observations of most 3D points.
Further, light fields do not entail any form of scene geometry, which allows them to capture complex view dependence but poses significant challenges in interpolating unseen rays in a geometrically meaningful way.
Existing methods address these issues by sacrificing view interpolation~\cite{feng2021signet}, leveraging data driven priors~\cite{sitzmann2021light}, or relying on strong supervision signals~\cite{lindell2021autoint}.

In order to deal with these challenges, we employ a novel ray-space embedding network that re-maps the input ray-space into an embedded latent space. This facilitates both the registration of rays observing same 3D points and the interpolation of unobserved rays, which leads to better memorization and view synthesis at the same time.
The embedding network alone already provides state-of-the-art view synthesis quality for densely sampled inputs (such as the Stanford light fields~\cite{wilburn2005high}). However, it does not interpolate well in sparser input sequences (such as those from Real Forward-Facing~\cite{mildenhall2020nerf}).
%
%
Thus, we represent such scenes with a set of local light fields, where each local light field is less prone to large depth and texture changes.
Each local light field has to learn a simpler embedding at the price of several more network evaluations per ray.

We evaluate our method for learning neural light fields in sparse and dense regimes, with and without subdivisions. We compare to state-of-the-art view-synthesis methods and show that our approach achieves comparable or better view synthesis quality in both regimes, in a fraction of render time that other methods require, while still maintaining their small memory footprint (Figure~\ref{fig:teaser}).

In summary, our contributions are:
\begin{enumerate}[leftmargin=*,topsep=5pt]
    \item A novel neural light field representation that employs a ray-space embedding network and achieves state-of-the-art quality for small-baseline view synthesis without any geometric constraints.
    \item A subdivided neural light field representation for large baseline light fields that leads to a good trade-off in terms of the number of network queries vs.\ quality, which can be optimized to achieve comparable performance to NeRF \cite{mildenhall2020nerf} and NeX \cite{wizadwongsa2021nex} for sparse real-world scenes.
    \item Improved capture of view-dependent appearance in both sparse and dense regimes (e.g., complicated reflections and refractions) that existing volume-based methods such as NeRF~\cite{mildenhall2020nerf} and NeX~\cite{wizadwongsa2021nex} struggle to represent.
\end{enumerate}
\section{Related Work}

\noindent
\topic{Light Fields}
Light fields~\cite{levoy1996light} or Lumigraphs~\cite{gortler1996lumigraph} represent the radiance entering all positions in 3D space, in all viewing directions.
By resampling the rays from the discrete set of captured images, one can query arbitrary rays in a light field and synthesize photorealistic novel views of a static scene without necessitating geometric reconstruction. While light fields have the flexibility to represent complicated, non-Lambertian appearance, this often requires high sampling rates, and thus capturing and storing an excessive number of images.
Much effort has been devoted to extending light fields to sparse, unstructured input sequences~\cite{buehler2001unstructured,davis2012unstructured} and synthesizing (i.e. interpolating) in-between views for such sequences~\cite{shi2014light,kalantari2016learning,vagharshakyan2017light,wu2017light,yeung2018fast,wu2019learning}.
Very recently, neural implicit representations have been applied to modeling light fields for better memory compactness~\cite{feng2021signet} and faster rendering for view synthesis~\cite{sitzmann2021light,lindell2021autoint}.
Like these methods, our work leverages neural light fields as its core representation for view synthesis.
Unlike prior work, which sacrifices view interpolation~\cite{feng2021signet} or leverages meta-learning for strong data-driven priors~\cite{sitzmann2021light}, we enable view synthesis without explicit priors, using per-scene training only.
%
%

\topic{Representations for View Synthesis}
Various 3D representations have been developed for rendering novel views from a sparse set of captured images.
Image-based rendering techniques leverage proxy geometry to warp and blend source image content into novel views~\cite{shum2000review,buehler2001unstructured,chaurasia2013depth,penner2017soft}. 
Recent advances in image-based rendering include learning-based disparity estimation~\cite{Flynn_2016_CVPR,kalantari2016learning}, blending~\cite{hedman2018deep, Riegler2020FVS}, and image synthesis via 2D CNNs ~\cite{Riegler2020FVS,Riegler2021SVS}.
Other commonly used scene representations include voxels~\cite{sitzmann2019deepvoxels,lombardi2019neural}, 3D point clouds~\cite{aliev2020neural,ruckert2021adop,kopanas2021point} or camera-centric layered 3D representations, such as multi-plane images~\cite{zhou2018stereo,mildenhall2019local,broxton2020immersive,tucker2020single,wizadwongsa2021nex} 
or layered depth images~\cite{hedman2017casual,tulsiani2018layer,kopf2020one,shih20203d}. 
Rather than using discrete representations as in the above methods, a recent line of work explores neural representations to encode the scene geometry and appearance as a continuous volumetric field~\cite{sitzmann2019scene,bemana2020xfields,mildenhall2020nerf,liu2020nsvf,martin2020nerf}.
We refer the readers to \cite{tewari2020state} for more comprehensive discussions of these works. 

Many of these methods learn a mapping from 3D points and 2D viewing directions to color/density, which can be rendered using numerical integration -- thus, rendering the color of a pixel for novel views requires hundreds of MLP evaluations.
Instead, our approach directly learns the mapping from rays to color, and therefore supports efficient rendering with fewer network evaluations.
In some ways, our method is similar to AutoInt~\cite{lindell2021autoint}, which predicts integrated radiance along ray segments and uses subdivision.
The accuracy of AutoInt, however, is lower than NeRF due to the approximation of the nested volume rendering integral.
By contrast, our approach leads to comparable performance to NeRF.

\topic{Coordinate-based Representations}
Coordinate-based representations have emerged as a powerful tool for overcoming the limitations of traditional discrete representations (e.g., images, meshes, voxelized volumes).
The core idea is to train an MLP to map an input coordinate to the desired target value such as pixel color~\cite{sitzmann2020implicit,xiang2021neutex,oechsle2020learning,martel2021acorn}, signed distance~\cite{park2019deepsdf,chabra2020deep}, occupancy~\cite{mescheder2019occupancy}, volume density~\cite{mildenhall2020nerf}, or semantic labels~\cite{Zhi2021inplace}.
Like existing coordinate-based representation approaches, our method also learns the mapping from an input coordinate (ray) to a target scene property (color).
Our core contributions lie in designing a coordinate embedding network that enables plausible view synthesis.

\topic{Learned Coordinate Embedding}
Learned coordinate transformation or embedding has been applied to extend the capability of coordinate-based representations, such as NeRF, for handling higher dimensional interpolation problems. 
For example, several works tackle dynamic scene view synthesis~\cite{li2020neural,park2021nerfies,xian2021space,gao2021dynamic,tretschk2021non} by mapping space at each time-step into a canonical frame. 
Others leverage coordinate embedding for modeling articulated objects such as the human body~\cite{liu2021neural,peng2021animatable}.
Our ray-space embedding network can be viewed as locally deforming the input ray coordinates such that the MLP can produce plausible view interpolation while preserving faithful reconstruction of the source views.

\topic{Subdivision}
Recent coordinate-based representations employ 3D space partitioning/subdivision either for improving the rendering efficiency~\cite{yu2021plenoctrees, reiser2021kilonerf,hedman2021baking} or representation accuracy~\cite{rebain2021derf,martel2021acorn}.
Our work also leverages spatial subdivision and shows improved rendering quality, particularly on scenes with large camera baselines.

\section{Neural Light Fields}

\noindent
A neural radiance field~\cite{mildenhall2020nerf} represents the appearance and geometry of a scene with an MLP
$\network : \left(\xpos_t, \dirout\right) \rightarrow \left( L_\text{e}\left(\xpos_t, \dirout\right), \sigma(\xpos_t) \right)$ with trainable weights $\boldsymbol\theta$. It takes as input a 3D position $\xpos_t$ and a viewing direction $\dirout$, and produces both the density $\sigma(\xpos_t)$ at point $\xpos_t$ and the radiance $L_\text{e}(\xpos_t,\dirout)$ emitted at point $\xpos_t$ in direction $\dirout$.

One can generate views of the scene from this MLP using volume rendering:
\begin{equation}
    L(\xpos, \dirout) = \int_\tnear^\tfar \!\!\!
    \underbrace{T\left(\xpos, \xpos_t\right)}_\text{\scriptsize Acc. transmittance} \,
    \underbrace{\sigma\left(\xpos_t\right)}_\text{\scriptsize Density} \,
    \underbrace{L_\text{e}\left(\xpos_t, \dirout\right)}_\text{\scriptsize Emitted radiance} \, \mathrm{d}t,
\label{eqn:volume_rendering}
\end{equation}
where 
$T\left(\xpos, \xpos_t\right)=
e^{- \int_{\,t_n}^t
\sigma( \xpos_s) \,
\mathrm{d}s}
$ describes the accumulated transmittance for light propagating from position $\xpos$ to $\xpos_t$, for near and far bounds $t \in \left[\tnear, \tfar\right]$ of the scene.
In practice, the integral is approximated using numerical quadrature:
\begin{equation}
    L(\xpos, \dirout) \approx \sum_{k = 1}^{N} \hat{T}\left(\xpos, \xpos_k\right) (1 - e^{-\sigma(\xpos_k) \Delta\xpos_k}) \,
    L_\text{e}\left(\xpos_k, \dirout \right),
\label{eqn:quadrature}
\end{equation}
where 
$\hat{T}(\xpos, \xpos_k)=e^{ - \sum_{j=1}^k \sigma(\xpos_j)\Delta\xpos_j}$ and 
$\Delta\xpos_j = \xpos_j - \xpos_{j-1}$ is the distance between adjacent samples. 
An accurate approximation of (\ref{eqn:volume_rendering}) requires many samples and thus many neural network evaluations (on the order of hundreds) per ray.

\label{sec:neural_light_field_baseline}

\topic{Light Fields}
Whereas a radiance field represents the radiance emitted at each point in space, a light field represents the total integrated radiance traveling along a ray.
In other words, Equations (\ref{eqn:volume_rendering}) and (\ref{eqn:quadrature}) describe the relationship between the light field $L$ on the left-hand side and the radiance field $(L_e, \sigma)$ on the right-hand side.
By optimizing a neural light field representation for $L$, given an input ray, one can predict this integrated radiance (and therefore render the color along this ray) with only a single neural network evaluation.

Assuming that we are operating on forward facing scenes, and that radiance remains constant along rays,
we can construct a 4D ray space by intersecting rays with two planes, $\firstplane$ and $\secondplane$.
We can then take the local plane coordinates, $(x, y) \in \firstplane$ and $(u, v) \in \secondplane$ of these ray intersections as the ray parameterization for the 4D light field, $L: \ray = (\rayparam) \rightarrow \col = (r, g, b)$.

\begin{figure}[t]
\centering
\small
\includegraphics[width=.9\linewidth]{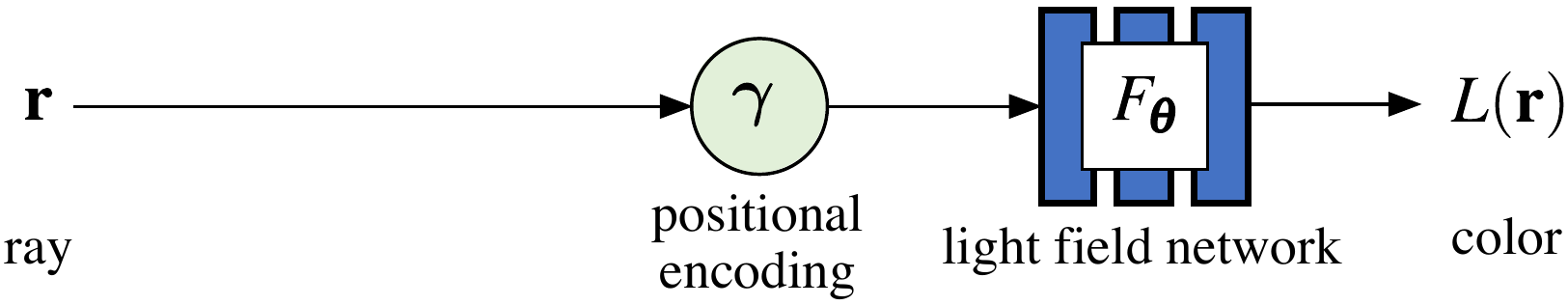}\\
(a) Baseline (without ray-space embedding)\\[3mm]
\includegraphics[width=.9\linewidth]{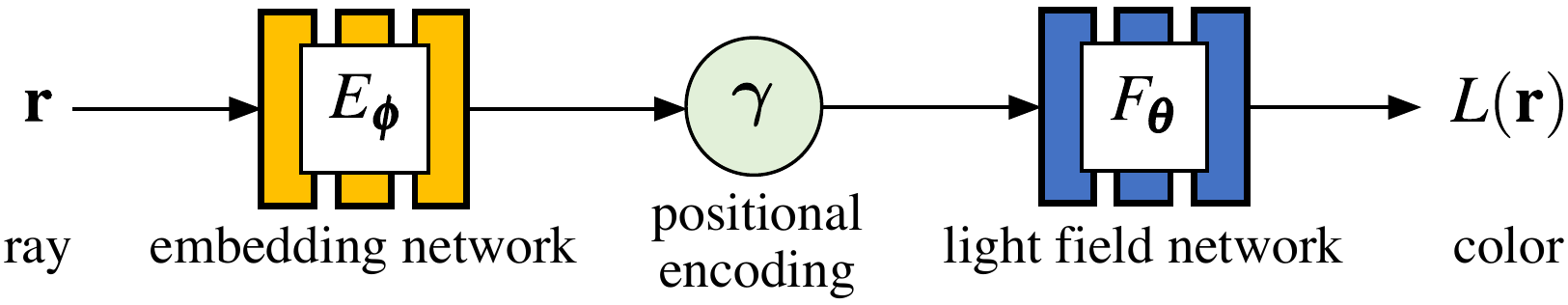}\\
(b) Feature-based embedding\\[3mm]
\includegraphics[width=.9\linewidth]{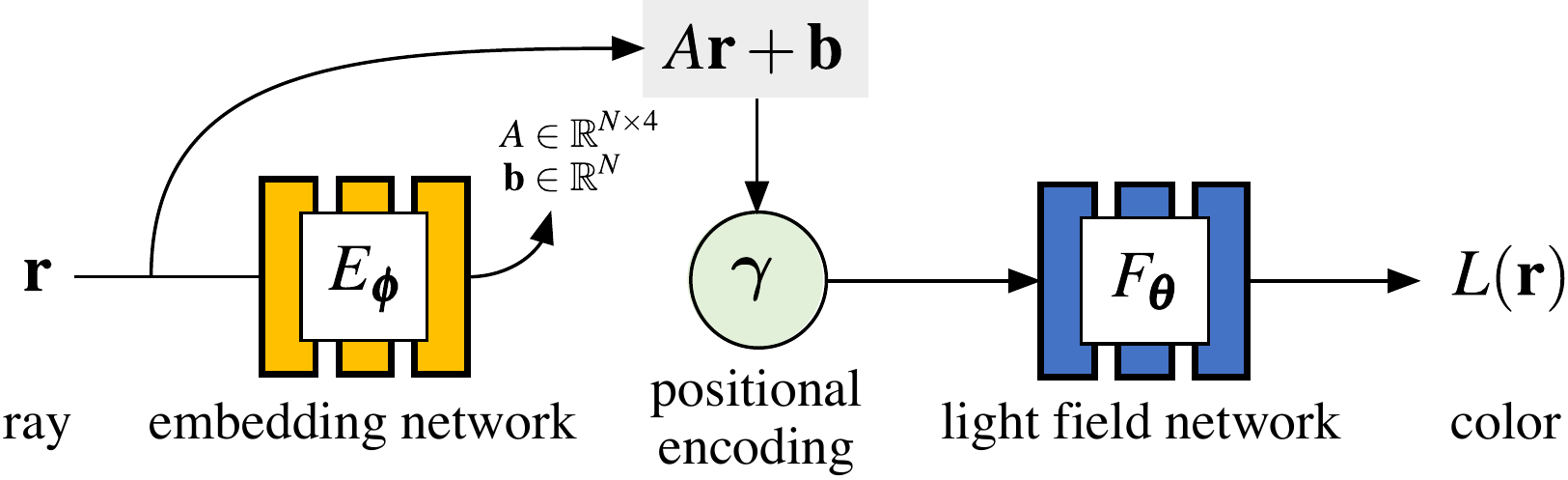}\\
(c) Local affine transformation--based embedding
\caption{
\tb{Ray-space embedding networks}. 
Building upon the baseline (a), we explore two different schemes for ray-space embedding: (b) predicting the latent features and (c) predicting affine transformation parameters. 
}
\vspace{-5mm}
\label{fig:embedding}
\end{figure}

\topic{Baseline Neural Light Fields}
To begin with, similar to how NeRF uses an MLP to represent radiance fields, we define an MLP $F_{\boldsymbol\theta}$ to represent a light field. It takes as input positionally encoded 4D ray coordinates $\posenc(\ray)$ and outputs  color $\col$ (integrated radiance) along each ray (Figure~\ref{fig:embedding}a):
\begin{equation}
\label{eq:baseline}
    L_\mathrm{base}(\ray) = F_{\boldsymbol\theta} (\posenc(\ray)) \, .
\end{equation}
Unfortunately, this baseline approach is an unsatisfactory light field representation due to the following challenges.
First, the captured input images only provide partial observations in the form of a sparse set of 2D slices of the full 4D ray space, so that each 4D ray coordinate in the input training data is observed at most once, assuming that no cameras share the same origin.

Second, light fields do not explicitly represent 3D scene geometry; hence the network a priori does not know how to interpolate the colors of unobserved rays from training observations.
In other words, when querying a neural network representing a light field with unseen ray coordinates, multi-view consistency is not guaranteed.
To address these challenges, we present two key techniques --- ray-space embedding and subdivision --- which substantially improve the rendering quality of the proposed neural light field representation.

\section{Ray-Space Embedding Networks}
\label{sec:ray_embedding}

\noindent
We first introduce \emph{ray-space embedding networks}. 
These networks re-map the input ray-space into an embedded latent space. 
Intuitively, our goals for ray-space embedding networks are:
\begin{enumerate}[leftmargin=*,topsep=5pt]
\item \textbf{Memorization:} Map disparate coordinates in input 4D ray-space that observe the same 3D point to the same location in the latent space. This allows for more stable training and better allocation of network capacity, and thus better memorization of input views.
\item \textbf{Interpolation:} Produce an interpolable latent space, such that querying unobserved rays preserves multi-view consistency and improves view synthesis quality.
\end{enumerate}

\topic{Feature-Space Embedding}
A straightforward approach would be to learn a nonlinear mapping from 4D ray coordinates to a latent feature space via an MLP,
$E^\mathrm{feat}_{\boldsymbol\phi} : \ray \rightarrow \mathbf{f} \in \mathbb{R}^N$, where $N$ is the feature space dimension (see Figure~\ref{fig:embedding}b). 
This embedding network can produce a nonlinear, many-to-one mapping from distinct ray coordinates into shared latent features, which would allow for better allocation of capacity in the downstream light field network $F_{\boldsymbol\theta}$ . The finite capacity of $F_{\boldsymbol\theta}$ incentivizes this ``compression'' and thus encourages rays with similar colors to be mapped to nearby features in the latent space; effectively a form of implicit triangulation.

After embedding, we then feed the positionally encoded $N$-dimensional feature into the light field MLP $F_{\boldsymbol\theta}$:
\begin{equation}
\label{eq:feature}
    L_\mathrm{feat}(\ray) = F_{\boldsymbol\theta}(\posenc(E^\mathrm{feat}_{\boldsymbol\phi}(\ray))) \, .
\end{equation}
While this approach compresses and helps to find correspondences in ray space, it is not enough to facilitate the interpolation between these correspondences (see Section~\ref{sec:experiments}).

\begin{figure}[t]
\centering
\mpage{0.49}{%
    \includegraphics[width=\linewidth]{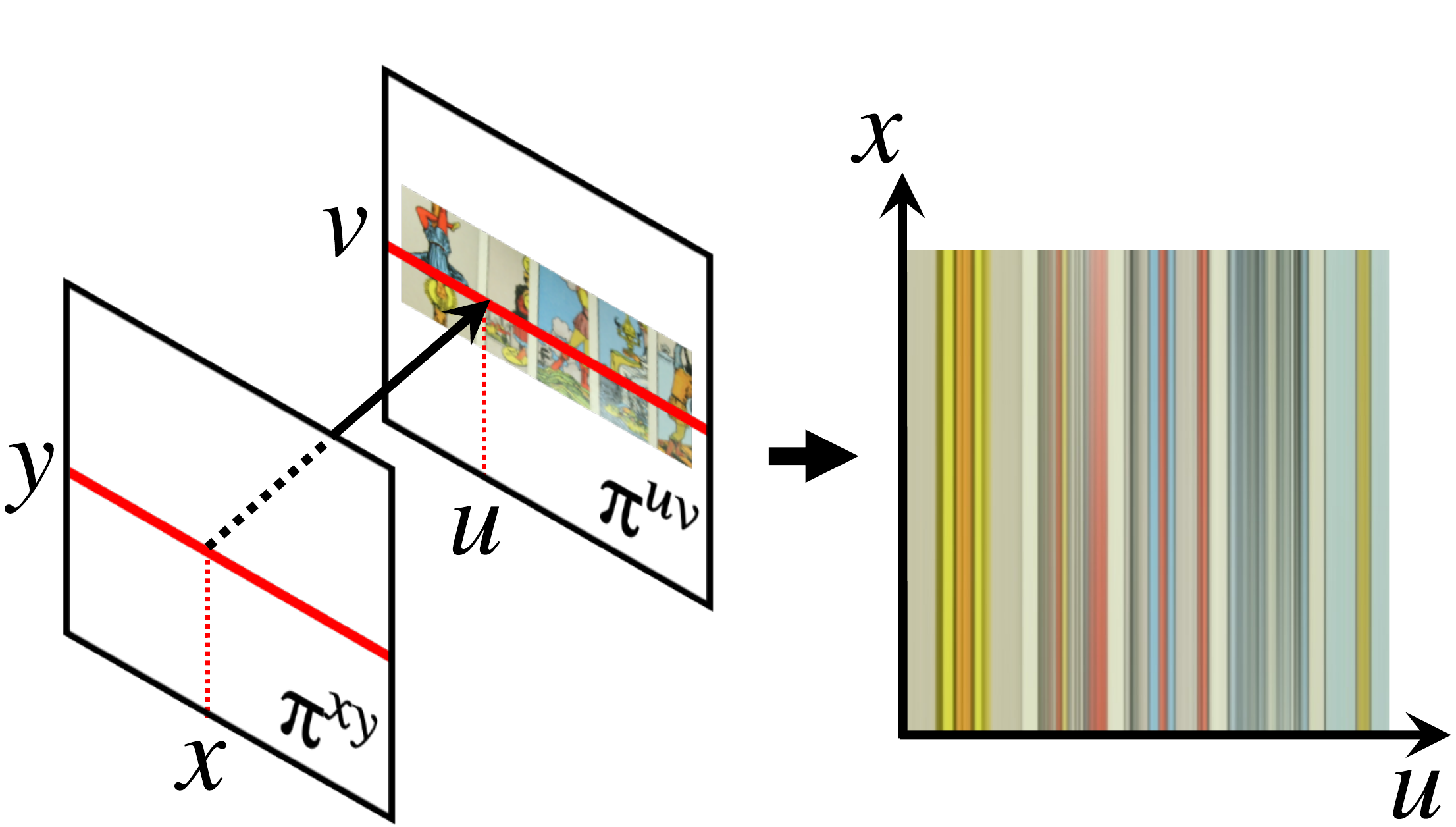}
	\footnotesize{(a) Axis-aligned ray-space}
}\hfill%
\mpage{0.49}{%
    \includegraphics[width=\linewidth]{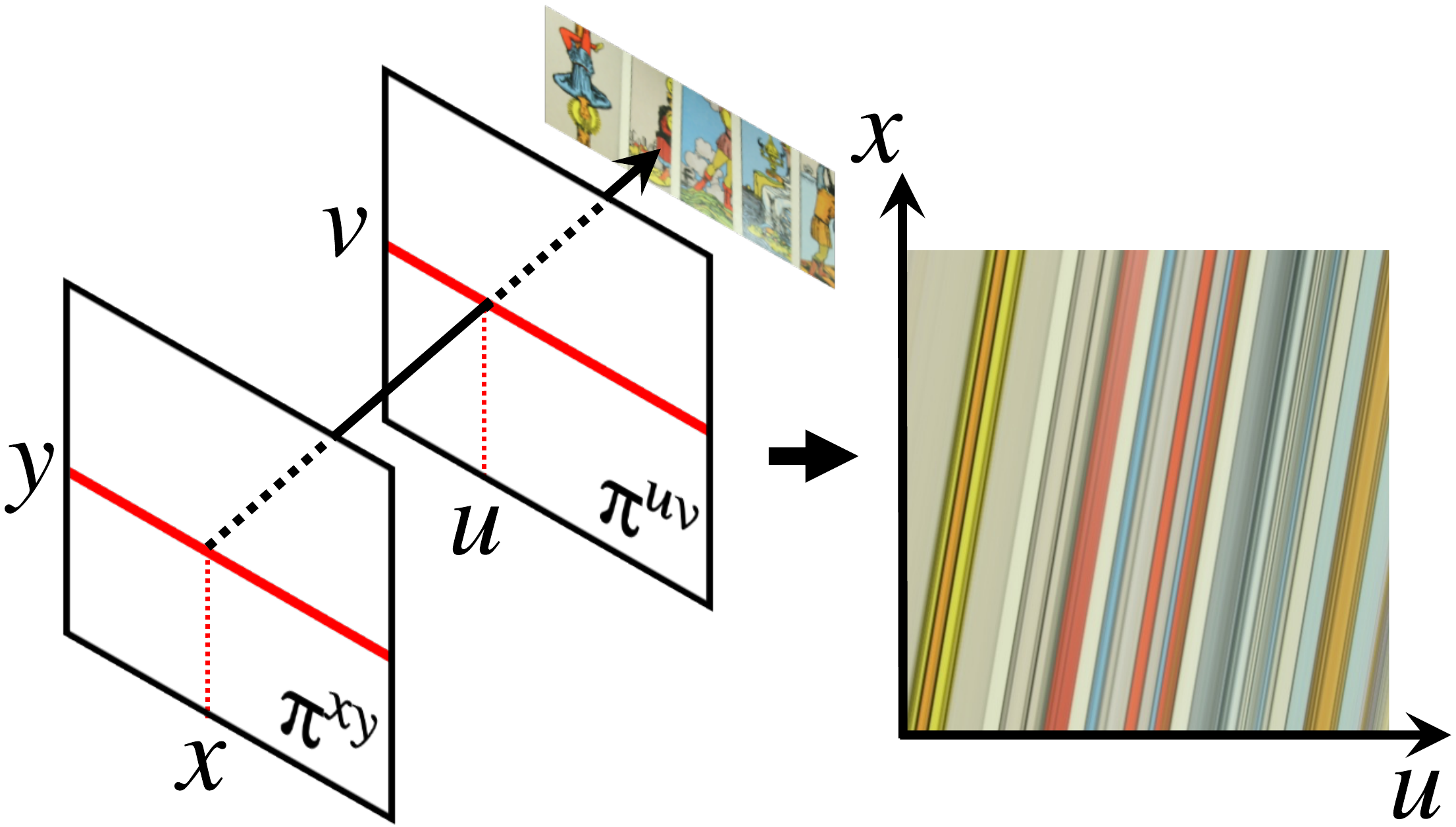}
	\footnotesize{(b) Ill-parameterized ray-space}
}%
\caption{
\tb{Importance of light field parameterization}. (a) It is easy to learn a light field for a well-parameterized ray space, but (b) a ray space with non-axis-aligned level sets makes the problem more challenging \cite{tancik2020fourier}. See Section~\ref{sec:ray_embedding} and appendix for discussions.
}
\label{fig:affine_motivation}
\vspace{-5mm}
\end{figure}

\topic{Light Field Re-parameterization}
To understand our goals better, consider a scene containing an axis-aligned textured square at a fixed $z$-depth.
Again, we assume a two-plane parameterization for the input ray-space of this scene. If one of the planes is at the depth of the square, then the light field will only vary in two coordinates, as illustrated in Figure~\ref{fig:affine_motivation}a.
In other words, the light field is only a function of $(u, v)$ and is constant in $(x, y)$.
Since positional encoding is applied separately for each input coordinate, the baseline model~\eqref{eq:baseline} can interpolate the light field perfectly if it chooses to ignore the positionally encoded coordinates $(x, y)$.

On the other hand, if the two planes in the parameterization do not align with the textured square, the color level sets (e.g., the line structures in the $ux$ slices) of the light field are 2D affine subspaces that are not axis-aligned (see Figure~\ref{fig:affine_motivation}b).
Axis-aligned positional encoding, as used by NeRF \cite{mildenhall2020nerf}, will yield interpolation kernels \cite{tancik2020fourier} that are ill-suited for capturing these sheared subspaces, especially from incomplete observations. One can therefore frame the task of effective interpolation as learning an optimal re-parameterization of the light field for a scene, such that the 2D color level sets for each 3D scene point are axis-aligned. See the appendix for a more extensive discussion.


\topic{Local Affine-Transformation Embedding}
Given this intuition, we describe our architecture for ray-space embedding, which learns local affine transformations for each coordinate in ray space.
We learn local affine transformations rather than a single global transformation because different rays may correspond to points with different depths, which will cause the shape of color level sets to vary.

To this end, we employ an MLP  $E_{\boldsymbol\phi} : \ray \rightarrow (A \in \mathbb{R}^{N \times 4}, \mathbf{b} \in \mathbb{R}^N)$.
The output of the network is a $N \times 4$ matrix $A$, as well as an $N$-vector $\mathbf{b}$ representing a bias (translation), which together form a 4D$\rightarrow$ $N$D affine transformation.
This affine transformation is applied to the input ray coordinate $\ray$ before being positionally encoded and passed into the light field network $F_{\boldsymbol\theta}$:
\begin{equation}
\label{eq:affine}
    L(\ray) = F_{\boldsymbol\theta} (\posenc(A \ray + \mathbf{b})) , \; \text{where} \; (A, \mathbf{b}) = E_{\boldsymbol\phi} (\ray) .
\end{equation}
See Figure~\ref{fig:embedding}c for an illustration of this model. 

While setting $N = 4$ above is perfectly reasonable, we use $N = 32$ in practice, which can be seen as providing multiple possible re-parameterizations per ray. 
Also note that learning arbitrary affine transforms, rather than a single $z$-depth for each ray, allows the network to better capture both \textit{angular frequencies} in the light field (due to object depth), as well as \textit{spatial frequencies} (due to object texture).

\section{Subdivided Neural Light Fields}
\label{sec:subdivision}

\noindent
Although this approach works well for dense light fields, our embedding network struggles to find long-range correspondences between training rays when training data is too sparse.
Moreover, even if it can discover correspondences, interpolation for unobserved rays in between these correspondences remains underconstrained.

\begin{figure}[t]
\centering
\includegraphics[width=.9\linewidth]{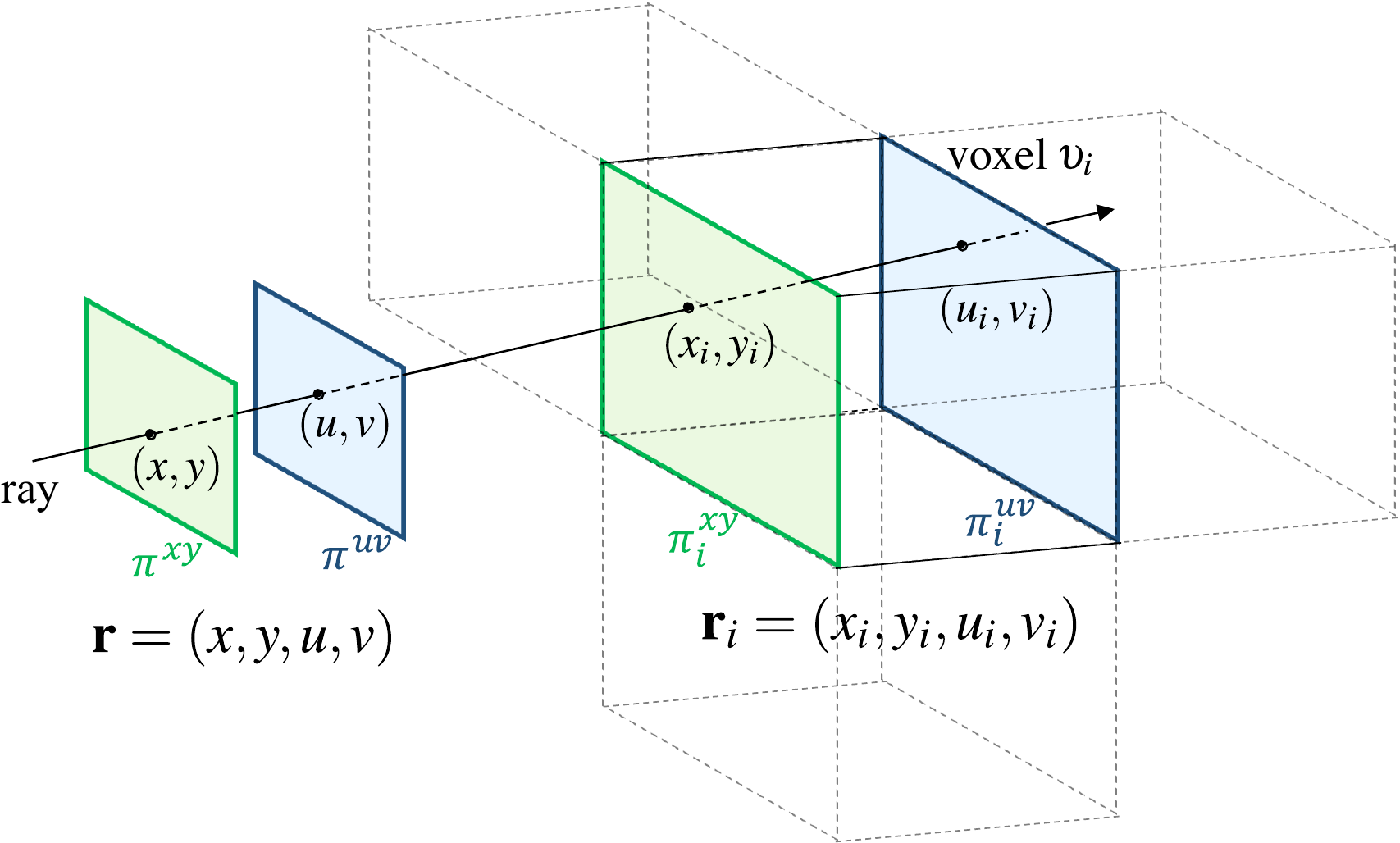}
\caption{
\tb{Re-parameterization for local light fields}. A ray $\ray = (x, y, u, v)$ defined globally is re-parameterized with respect to each local light field in voxel $\vox_i$, by intersecting the front and back faces of $\vox_i$. This yields local ray coordinates $\ray_i = (x_i, y_i, u_i, v_i)$.
}
\label{fig:ray_para}
\vspace{-5mm}
\end{figure}

To resolve these issues, we propose learning a voxel grid of local light fields that covers the entire 3D scene.
This approach is motivated by the following observation: if we parameterize a local light field for a voxel by intersecting rays with the voxel's front and back planes (See Figure~\ref{fig:ray_para}), the color level sets of the local light field are already almost axis-aligned.
As such, the ray-space embedding for each voxel can be simple, only inducing small shears of the input ray-space to allow for good interpolation. On the other hand, a global light field requires learning a complex ray-space embedding that captures occlusions and significant depth changes.

Therefore a local light field can be learned much more easily than an entire global light field with the same training data. 
This still requires that we solve an assignment problem:
we must know which rays to use to train each local light field.
While we can exclude all rays that do not intersect the voxel containing the local light field, many other rays may be occluded before they hit the voxel. Therefore, these rays should also be excluded during training.

A simple way to handle ray assignment is by learning (per-ray) opacity.
If the opacity accumulated along a ray before it hits a voxel is high, the ray should receive a little contribution from this voxel.
We, therefore, modify our light field network also to produce integrated opacity, or alpha for each ray. It is important to note that the opacity can vary depending on which ray is queried through a particular voxel --- opacity is a function of the ray, not the voxel itself.

\topic{Subdivided Volume Rendering}
We detail how a set of local light fields can be rendered to form new images.
Given a voxel grid in 3D space, we place a local light field within each voxel. A voxel's light field is parameterized with respect to its front and back planes (See Figure~\ref{fig:ray_para}), yielding ray coordinates $\ray_i$, where $i$ is some voxel index, for any ray $\ray$ defined globally.
If 3D space is subdivided into $M$ voxels, there exist $M$ distinct parameterizations $\{\ray_i\}_{i=1}^{M}$ of each $\ray$.

To implement this model, the embedding network $E_{\boldsymbol\phi}$ is augmented such that it takes the positionally encoded voxel index $\posenc(i)$ in addition to the 4D ray coordinates, i.e., $E_{\boldsymbol\phi} : (\ray_i, \posenc(i)) \rightarrow (A_i, \mathbf{b}_i)$. The light field network $F_{\boldsymbol\theta}$ is augmented in the same way and predicts both color \emph{and} alpha:
$F_{\boldsymbol\theta} : (\posenc(A_i \ray_i + \mathbf{b}_i), \posenc(i)) \rightarrow (\col_i, \alpha_i)$. Again, we emphasize that $\alpha_i$ is a function of the embedded ray, not just the voxel.

\begin{figure}[t]
\centering
\includegraphics[width=\linewidth]{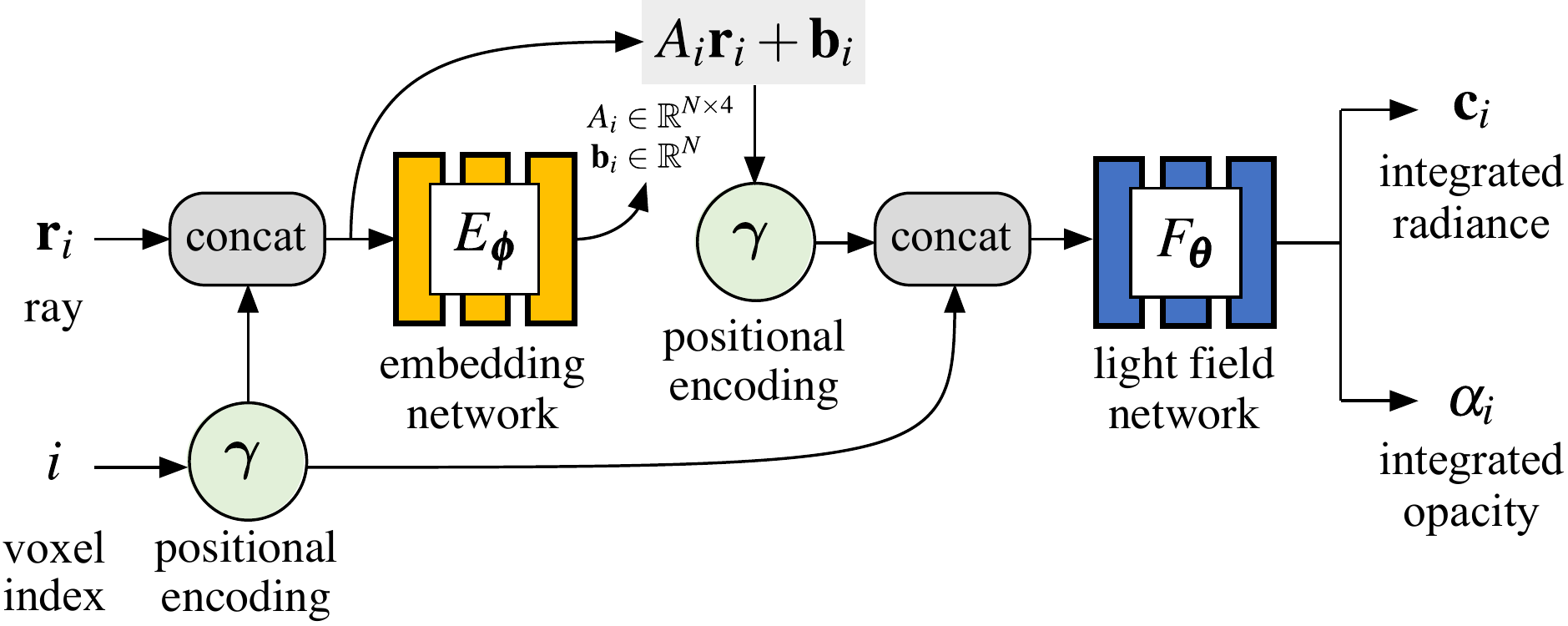}
\caption{
\tb{Local light fields via spatial subdivision}.
Each local light field takes a locally re-parameterized ray $\ray_i$ and its voxel index $i$, and predicts the radiance and opacity \emph{both} integrated for the ray segment traveling through the voxel. Both the embedding network $E_{\bm{\phi}}$ and the light field network $F_{\bm{\theta}}$ are conditioned on positionally encoded voxel index $\posenc(i)$.
}
\label{fig:subdivision}
\vspace{-2mm}
\end{figure}

Given this model, rendering works in the following way.
First, the set of voxels $\voxset (\ray)$ that the ray $\ray$ intersects are identified. The ray is then intersected with each voxel's front and back planes $\firstplane_i$ and $\secondplane_i$. This process yields a series of local ray coordinates $\{ \ray_i\}_{\vox_i \in \voxset(\ray)}$. 
The color $\col_i$ and alpha $\alpha_i$ of each $\ray_i$ are then computed as:
\begin{equation}
    (\col_i, \alpha_i)  = F_{\boldsymbol\theta} (\posenc(A_i \ray_i + \mathbf{b}_i), \posenc(i)) \, ,
\end{equation}
where the local affine-transformation is obtained from the embedding network:
\begin{equation}
    (A_i, \mathbf{b}_i) = E_{\boldsymbol\phi} (\ray_i, \posenc(i)) \, .
\end{equation}
The final color of ray $\ray$ is then over-composited:
\begin{equation}
    \col = \sum_{i \in \voxset(\ray)} \Big( \prod_{j \in \voxset(\ray) \wedge j < i} (1 - \alpha_j) \Big) \alpha_i \col_i \, ,
\end{equation}
which assumes that we sort the voxels by their distance from the ray origin in ascending order (see Figure~\ref{fig:subdivision}).


\section{Experiments}
\label{sec:experiments}

\noindent
\topic{Datasets}
We evaluate our model in two angular sampling regimes: sparse (wide-baseline) and dense (narrow-baseline), using the Real Forward-Facing dataset~\cite{mildenhall2020nerf}, the Shiny dataset~\cite{wizadwongsa2021nex}, and the Stanford Light Field dataset~\cite{wilburn2005high}.
Each scene from the Real Forward-Facing and Shiny datasets consist of a few dozen images with relatively large angular separation in between views.
We followed their authors' train/test split and hold out every 8th image from training. 

The Stanford light fields are all captured as  17$\times$17 dense 2D grids of images.
We take every 4th image in \emph{both} horizontal and vertical directions and use the resulting 5$\times$5 subsampled light field for training (thus using 25 images out of 289), while reserving the rest of the images for testing.

We downsample the images in the Real Forward-Facing and Shiny datasets to $504\times378$ pixels, and use half the original resolution of the Stanford images.
For quantitative metrics, we report PSNR, SSIM, LPIPS, and FPS; where FPS is computed for 512$\times$512 pixel images.

\begin{table}
\caption{\textbf{Quantitative comparisons.} We show the aggregated metrics over all scenes in each of the following datasets: Real Forward-Facing \cite{mildenhall2020nerf}, \textit{Undistorted} Real Forward-Facing \cite{wizadwongsa2021nex}, Shiny \cite{wizadwongsa2021nex}, and Stanford \cite{wilburn2005high}. }
\label{tab:quant}
\vspace{-2mm}
\renewcommand{\arraystretch}{1}
\begin{center}
\vspace{-1em}
\resizebox{\linewidth}{!}{%
\begin{tabular}{@{}llccc@{\hspace{0.75\tabcolsep}}r@{}}
  \toprule
  Dataset & Method & PSNR$\uparrow$ & SSIM$\uparrow$ & LPIPS$\downarrow$ & FPS$\uparrow$ \\
  \midrule
  \multirow{3}{*}{RFF}
  & NeRF~\cite{mildenhall2020nerf} & \textbf{27.928} & \textbf{0.916} & \underline{0.065} & 0.07 \\
  & AutoInt~\cite{lindell2021autoint} & 25.531 & 0.853 & 0.156 & \underline{0.26} \\
  & Ours (subdivision) & \underline{27.454} & \underline{0.905} & \textbf{0.060} & \textbf{0.34} \\
  \midrule
  \multirow{2}{*}{Undistorted RFF}
  & NeX~\cite{wizadwongsa2021nex} & \textbf{27.992} & \textbf{0.924} & \textbf{0.052} & 0.13 \\
  & Ours (subdivision) & 27.355 & 0.905 & 0.059 & \textbf{0.34} \\
  \midrule
    \multirow{2}{*}{Shiny}
  & NeX~\cite{wizadwongsa2021nex} & 28.268 & \textbf{0.940} & \textbf{0.036} & 0.13 \\
  & Ours (subdivision) & \textbf{28.499} & 0.931 & 0.038 & \textbf{0.34} \\
  \midrule
  \multirow{4}{*}{Stanford}
  & NeRF~\cite{mildenhall2020nerf} & \underline{37.559} & \underline{0.979}\phantom{0} & 0.037 &  0.07\\
  & NeX~\cite{wizadwongsa2021nex} & 36.286 & 0.972\phantom{0} & \underline{0.033} &  0.13\\
  & X-Fields~\cite{bemana2020xfields} & 35.559 & 0.976\phantom{0} & 0.036 & \textbf{20.00} \\
  & Ours (\emph{no} subdivision) & \textbf{38.054} & \textbf{0.982}\phantom{0} &  \textbf{0.020}  & \underline{9.14} \\
  \bottomrule
\end{tabular}%
}
\end{center}
\vspace{-8mm}
\end{table}

\topic{Baseline Methods}
We compare our model to NeRF~\cite{mildenhall2020nerf}, AutoInt~\cite{lindell2021autoint} (with 32 sections), and NeX~\cite{wizadwongsa2021nex} on the Real Forward-Facing dataset \cite{mildenhall2020nerf}; and additionally to NeX on the Shiny dataset \cite{wizadwongsa2021nex}.
As in \cite{wizadwongsa2021nex}, we perform evaluation on NeX before baking. 
More information about this is provided in the appendix.
In Table~\ref{tab:quant}, we include metrics for both the Real Forward-Facing and \textit{Undistorted} Real Forward-Facing dataset, a variant of the original dataset that is supported by the NeX codebase.
%
%
%
We did not include the following methods, although they are related, for various reasons: PlenOctrees~\cite{yu2021plenoctrees}, KiloNeRF~\cite{reiser2021kilonerf}, and NSVF~\cite{liu2020nsvf} operate on bounded 360-degree scenes only;
SNeRG~\cite{hedman2021baking} strictly underperforms NeRF on real forward-facing scenes.

On Stanford~\cite{wilburn2005high}, we compare our model to NeRF, NeX, and X-Fields~\cite{bemana2020xfields}, which is a state of the art method in image-based light field interpolation.
Kalantari et al.~\cite{kalantari2016learning} would also be a relevant comparison, but X-Fields outperforms it, hence we did not include it in our evaluations.

\topic{Subdivision}

For small-baseline light fields (Stanford \cite{wilburn2005high}), \emph{we do not use subdivision}, and use a single-evaluation per ray model with ray-space embedding. For sparse light fields (Real Forward Facing \cite{mildenhall2020nerf} and Shiny \cite{wizadwongsa2021nex}) we place local light fields within each voxel in a $32^3$ regular voxel grid that covers all of NDC space, with the exception of the CD and Lab sequences in the Shiny dataset, where we use a coarser $4^3$ voxel grid.

\topic{Implementation Details}
For our \emph{feature-based embedding}~\eqref{eq:feature}, we set the embedding space dimension to $N = 32$. The embedded feature $\mathbf{f}$ is $\ell_2$-normalized and then multiplied by $\sqrt{32}$, such that on average, each output feature has a square value close to one.
For our \emph{local affine transformation--based embedding}~\eqref{eq:affine}, we infer 32$\times$4 matrices and 32D bias vectors.
We then normalize the matrices $A$ with their Frobenius norm and multiply them by $\sqrt{32 \cdot 4}$. We use a \textit{tanh}-activation for our predicted bias vectors. 

For positional encoding $\posenc(\cdot)$ we use $L = 8$ frequency bands for our subdivided model and $L = 10$ frequency bands for our one-evaluation model. Higher frequencies are gradually eased in using windowed positional encoding, following Park et al.~\cite{park2021nerfies} --- over 50k iterations with subdivision, or 80k without.
For both the embedding and color models we use a $8$-layer, $256$ hidden unit MLP with one skip connection.
We share a single network each for all local light fields for both the embedding $E_{\boldsymbol\phi}$ and color $F_{\boldsymbol\theta}$ in our subdivided representation, which are ``indexed'' (see section \ref{sec:subdivision}) with the voxel center points $\mathbf{p}_i$, i.e., $i \equiv \mathbf{p}_i$.

We use a batch size of 4,096 for our subdivided model, and a batch size of 8,192 for our one-evaluation model during training and run all experiments on a single Tesla V100 GPU with 16\,GB of memory.
Please refer to the appendix and our project webpage for further details and results.
We will release the source code and pre-trained models to facilitate future research.

\section{Results}

\begin{figure}[t]
\footnotesize
\centering%
	\mpage{0.19}{%
		\begin{overpic}[width=\linewidth]{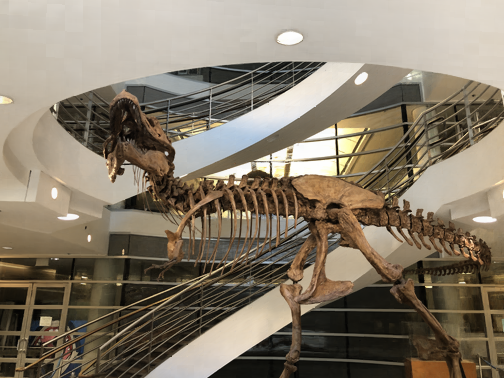}%
			\put(52, 1){\color{red}$\Box$}%
		\end{overpic}\\[-0.5mm]%
		\emph{T-Rex}
	}%
	\hspace{-1mm}\hfill%
	\mpage{0.8}{%
		\mpage{0.24}{%
			\includegraphics[width=\linewidth]{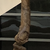}\\[0.2mm]%
		}\hspace{-1mm}\hfill%
		\mpage{0.24}{%
			\includegraphics[width=\linewidth]{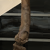}\\[0.2mm]%
		}\hspace{-1mm}\hfill%
		\mpage{0.24}{%
			\includegraphics[width=\linewidth]{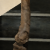}\\[0.2mm]%
		}\hspace{-1mm}\hfill%
		\mpage{0.24}{%
			\includegraphics[width=\linewidth]{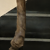}\\[0.2mm]%
		}
	}\\[1mm]%
	\mpage{0.19}{%
		\begin{overpic}[width=\linewidth]{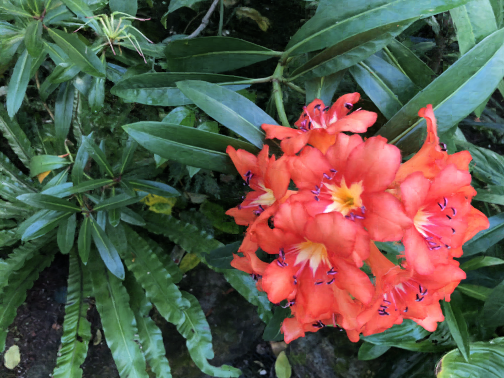}%
			\put(1, 63){\color{red}$\Box$}%
		\end{overpic}\\[-0.5mm]%
		\emph{Flower}
	}%
	\hspace{-1mm}\hfill%
	\mpage{0.8}{%
		\mpage{0.24}{%
			\includegraphics[width=\linewidth]{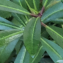}\\[0.2mm]%
		}\hspace{-1mm}\hfill%
		\mpage{0.24}{%
			\includegraphics[width=\linewidth]{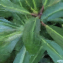}\\[0.2mm]%
		}\hspace{-1mm}\hfill%
		\mpage{0.24}{%
			\includegraphics[width=\linewidth]{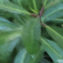}\\[0.2mm]%
		}\hspace{-1mm}\hfill%
		\mpage{0.24}{%
			\includegraphics[width=\linewidth]{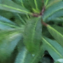}\\[0.2mm]%
		}
	}\\[1mm]%
	\mpage{0.19}{%
	}%
	\hspace{-1mm}\hfill%
	\mpage{0.8}{%
		\mpage{0.24}{%
			GT%
		}\hspace{-1mm}\hfill%
		\mpage{0.24}{%
			Ours%
		}\hspace{-1mm}\hfill%
		\mpage{0.24}{%
			NeRF~\cite{mildenhall2020nerf}%
		}\hspace{-1mm}\hfill%
		\mpage{0.24}{%
			NeX~\cite{wizadwongsa2021nex}%
		}
	}\\[2mm]%
\mpage{0.19}{%
		\begin{overpic}[width=\linewidth]{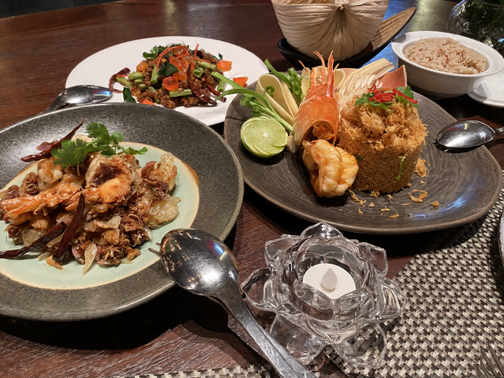}%
			\put(55, 20){\color{red}$\Box$}%
		\end{overpic}\\[-0.5mm]%
		\emph{Food}
	}%
	\hspace{-1mm}\hfill%
	\mpage{0.8}{%
		\mpage{0.32}{%
			\includegraphics[width=\linewidth]{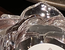}\\[0.2mm]%
		}\hspace{-1mm}\hfill%
		\mpage{0.32}{%
			\includegraphics[width=\linewidth]{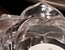}\\[0.2mm]%
		}\hspace{-1mm}\hfill%
		\mpage{0.32}{%
			\includegraphics[width=\linewidth]{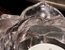}\\[0.2mm]%
		}
	}\\[1mm]%
\mpage{0.19}{%
		\begin{overpic}[width=\linewidth]{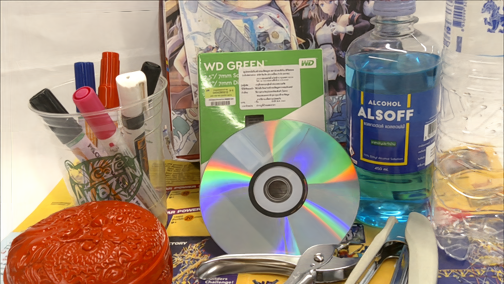}%
			\put(70, 15){\color{red}$\Box$}%
		\end{overpic}\\[-0.5mm]%
		\emph{CD}
	}%
	\hspace{-1mm}\hfill%
	\mpage{0.8}{%
		\mpage{0.32}{%
			\includegraphics[width=\linewidth]{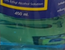}\\[0.2mm]%
		}\hspace{-1mm}\hfill%
		\mpage{0.32}{%
			\includegraphics[width=\linewidth]{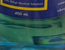}\\[0.2mm]%
		}\hspace{-1mm}\hfill%
		\mpage{0.32}{%
			\includegraphics[width=\linewidth]{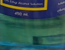}\\[0.2mm]%
		}
	}\\[1mm]%
	\mpage{0.19}{%
	}%
	\hspace{-1mm}\hfill%
	\mpage{0.8}{%
		\mpage{0.3}{%
			GT%
		}\hspace{-1mm}\hfill%
		\mpage{0.3}{%
			Ours%
		}\hspace{-1mm}\hfill%
		\mpage{0.3}{%
			NeX~\cite{wizadwongsa2021nex}%
		}
	}\\[2mm]%
	\mpage{0.19}{%
		\begin{overpic}[width=\linewidth,trim={0 1cm 0 0},clip]{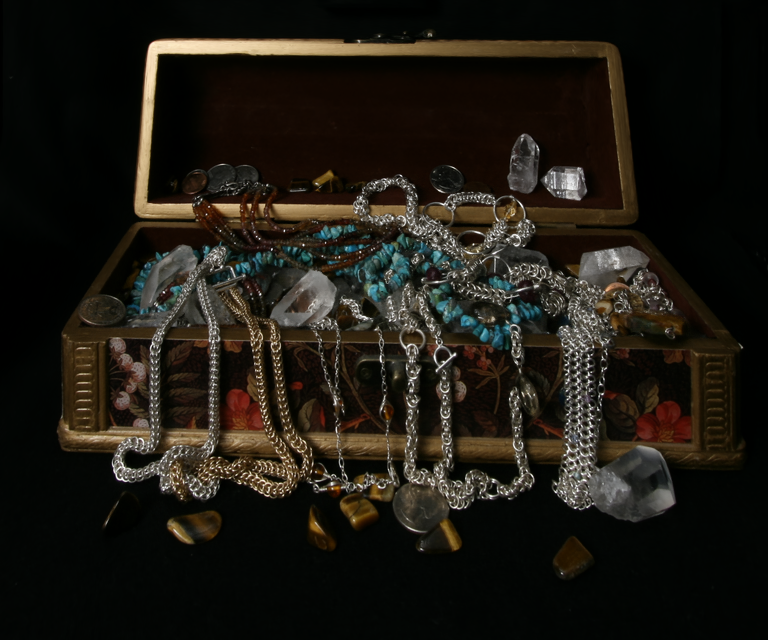}%
			\put(73.5, 39.5){\scriptsize\color{red}$\Box$}%
		\end{overpic}\\[-0.5mm]%
		\emph{Treasure}
	}%
	\hspace{-1mm}\hfill%
	\mpage{0.8}{%
		\mpage{0.24}{%
			\includegraphics[width=\linewidth]{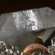}\\[0.2mm]%
		}\hspace{-1mm}\hfill%
		\mpage{0.24}{%
			\includegraphics[width=\linewidth]{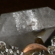}\\[0.2mm]%
		}\hspace{-1mm}\hfill%
		\mpage{0.24}{%
			\includegraphics[width=\linewidth]{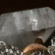}\\[0.2mm]%
		}\hspace{-1mm}\hfill%
		\mpage{0.24}{%
			\includegraphics[width=\linewidth]{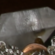}\\[0.2mm]%
		}%
	}\\[1mm]%
	\mpage{0.19}{%
		\begin{overpic}[width=\linewidth,trim={0 9cm 0 0},clip]{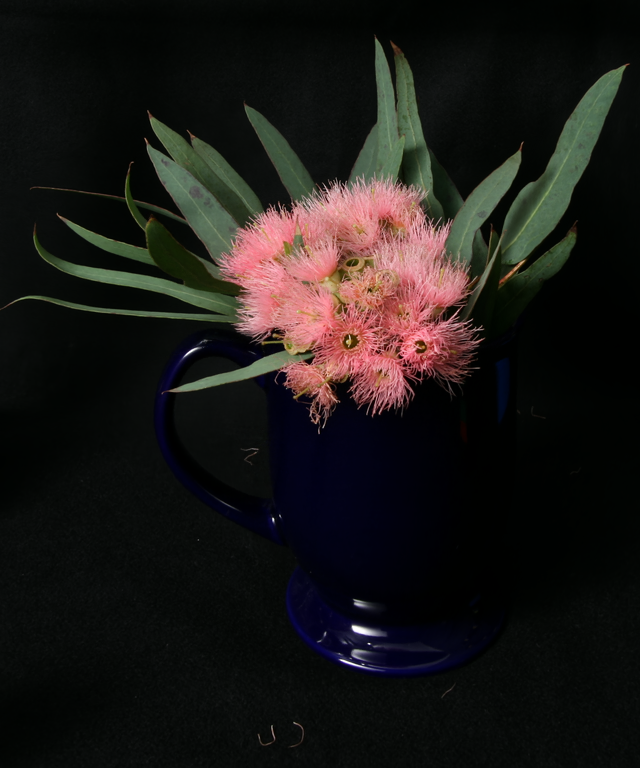}%
			\put(36.5, 37.5){\scriptsize\color{red}$\Box$}%
		\end{overpic}\\[-0.5mm]%
		\emph{Flowers}
	}%
	\hspace{-1mm}\hfill%
	\mpage{0.8}{%
		\mpage{0.24}{%
			\includegraphics[width=\linewidth]{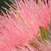}\\[0.2mm]%
		}\hspace{-1mm}\hfill%
		\mpage{0.24}{%
			\includegraphics[width=\linewidth]{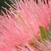}\\[0.2mm]%
		}\hspace{-1mm}\hfill%
		\mpage{0.24}{%
			\includegraphics[width=\linewidth]{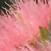}\\[0.2mm]%
		}\hspace{-1mm}\hfill%
		\mpage{0.24}{%
			\includegraphics[width=\linewidth]{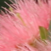}\\[0.2mm]%
		}%
	}\\[1mm]%
	\mpage{0.19}{%
	}%
	\hspace{-1mm}\hfill%
	\mpage{0.8}{%
		\mpage{0.24}{%
			GT%
		}\hspace{-1mm}\hfill%
		\mpage{0.24}{%
			Ours%
		}\hspace{-1mm}\hfill%
		\mpage{0.24}{%
			NeRF~\cite{mildenhall2020nerf}%
		}\hspace{-1mm}\hfill%
		\mpage{0.24}{%
			X-Fields~\cite{bemana2020xfields}%
		}%
	}\\[-2mm]%
\caption{%
\tb{Visual comparisons.} Close-ups reveal our representation leads to more faithful reconstruction of view-dependent effects, particularly distorted reflections and refractions (\emph{Horns} and \emph{T-Rex} from RFF; \emph{Treasure} from Stanford; and \emph{Food} and \emph{CD} from Shiny) as well as sharper details (\emph{Flower} from RFF and \emph{Flowers} from Stanford). See our project website for more results.
}%
\label{fig:qual}
\vspace{-5mm}
\end{figure}

\topic{Quantitative Evaluation}
Table~\ref{tab:quant} compares view synthesis quality using the metrics aggregated over held-out images in all scenes.
For the Real Forward-Facing dataset, our method outperforms AutoInt in both quality ($\sim$2\,dB) and speed ($\sim$30\%) while performing competitively with NeRF on PSNR and SSIM (within 0.5\,dB in quality) and rendering more than $\times$4.5 times faster. We also outperform NeRF on LPIPS. %
Additionally, we are competitive with NeX on the \emph{Undistorted} Real Forward-Facing dataset (within 0.65\,dB in quality). For the Shiny dataset, we outperform NeX in terms of PSNR, while performing slightly worse on SSIM and LPIPS metrics. We do especially well on the \emph{CD} and \emph{Lab} scenes with challenging view dependence, outperforming NeX by $>3$dB each. 

On Stanford, our method outperforms NeRF, NeX, and X-Fields in all three metrics.
Specifically, our model performs better on more complex scenes, e.g., \emph{Treasure} and \emph{Bulldozer}, with complicated geometry and high-frequency textures, while NeRF performs slightly better on simpler scenes like \emph{Chess} and \emph{Beans}. 
Our model renders a frame in about 0.11 seconds in this setup, which is faster than all other methods apart from X-Fields.

\topic{Training Time}
For both RFF and Shiny, it takes about 20 hours to train our model at $32^3$ resolution, while for the Stanford data, we train our model for about 10 hours. NeRF takes approximately 18 hours to train on all datasets, while NeX takes approximately $36$ hours on 2 GPUs. 
For X-Fields, we double the default capacity of their model as well as the number of images used for interpolation; and train X-Fields for 8 hours rather than the maximum time of $172$ minutes listed in their paper \cite{bemana2020xfields}.
See the appendix for more details.

\topic{Reconstruction of View Dependence}
Our approach reconstructs view dependence better than the baseline methods in both sparse and dense regimes (see Figures~\ref{fig:teaser}(\emph{left}) and~\ref{fig:qual}), due to the flexibility of the embedding network to predict different affine transforms for different regions of ray-space. This effectively produces locally deforming depth estimates, which can better model reflected and refracted content, for example. On the other hand, unlike our method, volumetric representations use a fixed global geometry, and are either late-conditioned on viewing direction \cite{mildenhall2020nerf}, or use lower frequency basis functions for view dependence \cite{wizadwongsa2021nex}. Thus, they struggle to represent high-frequency view dependent appearance that breaks multi-view color constraints.

In general, reproducing complex view dependence is a major challenge for volumetric representations, and is one of the primary contributions of our work. The difference is especially prominent in the Shiny dataset comparisons in Figure~\ref{fig:teaser}, as well as the results for \emph{T-rex}, \emph{Food}, \emph{CD}, and \emph{Treasure} in Figure~\ref{fig:qual}.

\topic{Memory Footprint}
Table~\ref{tab:memory} shows the model sizes.
Our model has a similar memory footprint to NeRF. NeX explicitly represents albedo for $16$ MPI layers, leading to larger memory consumption than pure coordinate-based neural representations.
%
%
Overall, our method provides the best trade-off between quality, rendering speed, and memory among the baselines we compared to; see the graphs in Figure~\ref{fig:teaser}(\emph{right}).

\begin{table}
\caption{\textbf{Memory footprint} of our model compared to other methods in terms of the number of trainable weights in each network. For X-Fields~\cite{bemana2020xfields} we include the memory footprint of the model, and of the training images which must be maintained in order to synthesize novel views.}
\label{tab:memory}
\vspace{-2mm}
\renewcommand{\arraystretch}{1}
\begin{center}
\vspace{-1em}
\resizebox{\linewidth}{!}{%
\begin{tabular}{@{}lccccc@{}}
  \toprule
  Method & \mbox{\quad Ours \quad} & NeRF~\cite{mildenhall2020nerf} & NeX \cite{wizadwongsa2021nex} & AutoInt~\cite{lindell2021autoint} & X-Fields~\cite{bemana2020xfields} \\
  \midrule
  Size (MB) & 5.4 & 4.6 & 38.4 & 4.6 & 3.8~/~75 \\
  \bottomrule
\end{tabular}%
}
\end{center}
\vspace{-3mm}
\end{table}

\begin{table}
\caption{\textbf{Ablations on the ray-embedding network} with subdivision (on Real Forward-Facing~\cite{mildenhall2020nerf}) and without subdivision (on Stanford~\cite{wilburn2005high} light fields).}
\label{tab:embedding}
\vspace{-2mm}
\renewcommand{\arraystretch}{1}
\renewcommand\cellalign{lc}
\begin{center}
\vspace{-1em}
\resizebox{\linewidth}{!}{%
\begin{tabular}{@{}lllccc@{}}
  \toprule
  Subdivision & Method & Embedding & PSNR$\uparrow$ & SSIM$\uparrow$ & LPIPS$\downarrow$ \\
  \midrule
  \multirow{4}{*}{\makecell{Used\\(On RFF)}}
  & \multirow{3}{*}{Ours}
  & Not used~\eqref{eq:baseline} & 26.696 & 0.888\phantom{0} & 0.079 \\
  && Feature~\eqref{eq:feature} & 26.922 & 0.891\phantom{0} & 0.073 \\
  && Local affine~\eqref{eq:affine} & \underline{27.454} & \underline{0.905}\phantom{0} & \textbf{0.060} \\
  \cmidrule(ll){2-6}
  & NeRF~\cite{mildenhall2020nerf} && \textbf{27.928} & \textbf{0.916}\phantom{0} & \underline{0.065} \\
  \midrule
  \multirow{4}{*}{\makecell{Not used\\(On Stanford)}}
  & \multirow{3}{*}{Ours}
  & Not used~\eqref{eq:baseline} & 25.111 & 0.841\phantom{0} & 0.098 \\
  && Feature~\eqref{eq:feature} & 37.120 & \underline{0.9792} & \underline{0.030} \\
  && Local affine~\eqref{eq:affine} & \textbf{38.054} & \textbf{0.982}\phantom{0} &  \textbf{0.020}  \\
  \cmidrule(ll){2-6}
  & NeRF~\cite{mildenhall2020nerf} && \underline{37.559} & 0.9790 & 0.037  \\
  \bottomrule
\end{tabular}%
}
\end{center}
\vspace{-5mm}
\end{table}

\topic{Ablations on Embedding Networks}
We study the effect of our proposed embedding networks via ablations.
See Table~\ref{tab:embedding} for quantitative evaluation with varying configurations.

With no embedding and only positional encoding (baseline), the network does not produce multi-view consistent view synthesis.
The use of our feature-based embedding leads to a boost in quality over the baseline. In particular, it achieves our first objective of ``memorization'' (identifying rays that observe the same 3D point), but still struggles with view interpolation. Our local affine transformation embedding achieves better interpolation quality because it implicitly encourages color level sets to be locally linear. Thus, we learn a model that interpolates in a (locally) view consistent way for free.

\topic{Ablations on Subdivision}
Table~\ref{tab:subdivision} and Figure~\ref{fig:teaser}(\emph{right}) summarize the trade-off between the quality and rendering speed of our model on various resolutions of spatial subdivision, compared to NeRF and AutoInt. For each subdivision level, we use a regular $N^3$ voxel grid, where $N \in (4, 8, 16, 32)$ 
As expected, our model archives better quality view synthesis as we employ more local light fields with finer spatial subdivision, trading off the rendering speed. The render-time increases linearly in $N$.
Our method provides better quality and speed than AutoInt at all comparable subdivision levels. 

\begin{table}
\caption{\textbf{Ablations on the subdivision} on sparse light fields (Real Forward-Facing ~\cite{mildenhall2020nerf}).}
\label{tab:subdivision}
\vspace{-2mm}
\renewcommand{\arraystretch}{1}
\begin{center}
\vspace{-1em}
\resizebox{\linewidth}{!}{%
\begin{tabular}{@{}llcccr@{}}
  \toprule
  Method & Subdivision \quad\quad & PSNR$\uparrow$ & SSIM$\uparrow$ & LPIPS$\downarrow$ & FPS$\uparrow$ \\
  \midrule
  \multirow{4}{*}{Ours}
  & $4^3$ grid & 25.579 & 0.867 & 0.094 & \textbf{2.82}\phantom{0} \\
  & $8^3$ grid & 26.860 & 0.893 & 0.070 & \underline{1.42}\phantom{0} \\
  & $16^3$ grid & 27.350 & 0.904 & \underline{0.062} & 0.69\phantom{0} \\
  & $32^3$ grid & \underline{27.454} & \underline{0.905} & \textbf{0.060} & 0.34\phantom{0} \\
  \midrule
  \multirow{3}{*}{AutoInt~\cite{lindell2021autoint} \quad}
  & 8 sections & 24.136 & 0.820 & 0.176 & 0.94\phantom{0} \\
  & 16 sections & 24.898 & 0.836 & 0.167 & 0.51\phantom{0} \\
  & 32 sections & 25.531 & 0.853 & 0.156 & 0.26\phantom{0} \\
  \midrule
  NeRF~\cite{mildenhall2020nerf}
  & & \textbf{27.928} & \textbf{0.916} & 0.065 & 0.07\phantom{0} \\
  \bottomrule
\end{tabular}%
}
\end{center}
\vspace{-7mm}
\end{table}

\topic{Limitations}
One limitation of our work is that we focus on light fields parameterized with two planes.
Although common, this does not encompass 360$^{\circ}$ scenes.
In the appendix, we show preliminary results for our method on 360$^{\circ}$ scenes using Pl\"ucker coordinates~\cite{sitzmann2021light}, which again outperforms AutoInt. However, color level sets for Pl\"ucker coordinates are no longer affine. Thus, we believe that improving the design of embedding networks for different parameterizations could lead to a larger boost in performance.

Additionally, our representation requires subdivision for sparse light fields. This comes at the cost of both increased render time and training time. We hope to investigate other network designs or regularization schemes, which encourage the representation to be multi-view consistent without subdivision.
Adaptive subdivision~\cite{liu2020nsvf,martel2021acorn} is another interesting direction for future work which could lead to better quality without sacrificing speed.

\section{Conclusion}

\noindent
We present a novel ray-space embedding approach for learning neural light fields, which achieves state-of-the-art quality on small-baseline datasets. 
To better handle sparse input, we leverage spatial subdivision with a voxel grid of local light fields, which improves quality at the cost of increased render time.
Our subdivided representation enables comparable performance to existing models, and achieves a better trade-off between quality, speed, and memory.
Additionally, in both regimes, our method can handle complex view dependent effects that existing state-of-the-art volumetric scene representations do not faithfully reproduce \cite{mildenhall2020nerf,wizadwongsa2021nex}. 

We believe that our approach is orthogonal to contemporary works that leverage hybrid or explicit grid-based representations for view synthesis\cite{yu2021plenoxels,muller2022instant,sun2021direct}, as well as approaches that use many small MLPs within a 3D volume \cite{reiser2021kilonerf}. As such, our approach can potentially be used to reduce the required grid resolution and number of sample points per ray for these methods --- which may further improve rendering quality, speed, and memory overhead .
We hope that our work can open up new avenues for view synthesis; and further, that its insights can be leveraged for other signal-representation problems (such as image/video representation) or high-dimensional interpolation problems (such as view-time or view-time-illumination interpolation~\cite{bemana2020xfields}).

{\small
\bibliographystyle{ieee_fullname}
\bibliography{main}
}

\appendix
\counterwithin{figure}{section}
\counterwithin{table}{section}

\section{Implementation Details}

\noindent
Our code is written entirely in python, using the PyTorch Lightning framework \cite{Falcon_PyTorch_Lightning_2019}. The design of our codebase was inspired by \cite{queianchen_nerf}. Below we include additional implementation details for our method for both the dense dataset (Stanford Light Field~\cite{wilburn2005high}) and sparse forward facing datasets (Real Forward-Facing~\cite{mildenhall2020nerf} and Shiny~\cite{wizadwongsa2021nex}). We also include information about the NeRF~\cite{mildenhall2020nerf}, NeX~\cite{wizadwongsa2021nex}, and AutoInt~\cite{lindell2021autoint} baselines.

\subsection{Stanford Light Field Dataset}
\label{sec:stanford_data}

\noindent
Calibration information is not provided with the Stanford Light Field dataset, apart from the $(x, y)$ positions of all images on the camera plane $\firstplane$. All Stanford Light Fields are parameterized with respect to a plane $\secondplane$ that approximately cuts through the center of each scene. Thus, a pixel coordinate in an image corresponds the location that a ray intersects $\secondplane$. We heuristically set the location of $\firstplane$ to $z = -1$, and the location of the object plane $\secondplane$ to $z = 0$. We scale the camera positions so that they lie between $[-0.25, 0.25]$ in both $x$ and $y$, and the pixel coordinates on $\secondplane$ so they lie between $[-1, 1]$. The camera positions, and the vector from the camera origin to the location of the pixel on the object plane then comprise our ray origins and directions.

For our method, we take camera coordinates and object plane coordinates, which correspond to intersections of the rays with $\firstplane$ and $\secondplane$, as our initial two-plane ray parameterization.

For NeRF/NeX, we set the near distance to $0.5$ and the far distance to $2$ for all scenes, except for the Knights scene, where we set the near distance to $0.25$. Note that, as defined, the scene coordinates may differ from the true (metric) world space coordinates by a projective transform. However, multi-view constraints still hold (intersecting rays remain intersecting, epipolar lines remain epipolar lines), and thus NeRF is able to learn an accurate volume.

\subsection{Real Forward-Facing Dataset}

\noindent
For the Real Forward-Facing Dataset, we perform all experiments in NDC space. For our subdivided model, we re-parameterize each ray $\ray$ within a voxel $\vox$ by first transforming space so that the voxel center lies at the origin. We then intersect the transformed ray with the voxel's front and back planes, and take the $xy$ coordinates of these intersections, which lie within the range $[-\textit{voxel width}, +\textit{voxel width}]$ in all dimensions as the parameterization.

\subsection{Shiny Dataset}

\noindent
Our procedure for evaluating on Shiny \cite{wizadwongsa2021nex} is identical to the Real Forward-Facing dataset. We perform experiments in NDC space and use a $32^3$ voxel grid that covers all of NDC space on all scenes \textit{except} CD/Lab where we use a coarser $4^3$ grid.

\subsection{NeRF and AutoInt Baselines}

\noindent
For the Stanford Dataset, we train NeRF for 400k iterations with a batch size of 1,024 for all scenes. For the Real Forward-Facing dataset, AutoInt does not provide pretrained models and training with their reported parameters on a V100 GPU with 16GB of memory leads to out-of-memory errors. They also do not provide multi-GPU training code. As such, we report the quantitative metrics for their method and for NeRF published in their paper, which come from models trained at the same resolution ($504\times378$), and using the same heldout views as ours on the Real Forward-Facing Dataset. 

\subsection{NeX Baseline}
\noindent We train NeX on all datasets (Stanford, Undistorted RFF, Shiny), using their public codebase, for $4000$ epochs on all scenes (the default number of epochs in their training script), or about $36$ hours each. We use their multi-GPU training code to split training over $2$ 16GB V100 GPUs. We note that NeX can perform real-time rendering after discretizing their view dependent basis functions into $400\times400$ textures. However the NeX codebase does not include evaluation code for their real-time renderable MPIs, and thus the numbers for NeX in Table 1 of the main paper are reported \emph{before} baking. While this presumably leads to an increase in quality, it takes a longer time to render images, hence the smaller FPS scores in Table 1. 

\section{Importance of Light Field Parameterization}

\noindent
Here, we expand on the discussion in Section 4 of the main paper and describe why our light field parameterization is crucial for enabling good view interpolation.
Let the two planes in the initial two plane parameterization be $\firstplane$ and $\secondplane$, with local coordinates $(x, y)$ and $(u, v)$.
In addition, let us denote the plane of the textured square as $\pi^{st}$ with local coordinates $(s, t)$, and assume that it is between $\firstplane$ and $\secondplane$.
Assume, without loss of generality, that the depth of $\firstplane$ is $0$, and suppose the depths of $\pi^{st}$ and $\secondplane$ are $z_{st}$ and $z_{uv}$ respectively.

For a ray originating at $(\hat{x}, \hat{y})$ on $\firstplane$ and passing through $(\hat{s}, \hat{t})$ on $\pi^{st}$, we can write (by similar triangles):
\begin{align}
    \hat{s} - \hat{x} &= (\hat{u} - \hat{x}) \frac{z_{st}}{z_{uv}} \, , \\
    \hat{t} - \hat{y} &= (\hat{v} - \hat{y}) \frac{z_{st}}{z_{uv}} \, ,
\end{align}
which gives
\begin{align}
    \hat{u} &= \frac{\hat{x} (z_{st} - z_{uv}) + \hat{s}}{z_{st}} \, , \\
    \hat{v} &= \frac{\hat{y} (z_{st} - z_{uv}) + \hat{t}}{z_{st}} \, .
\end{align}
Recall that the positional encoding of the 4D input parameterization $\gamma(\hat{x}, \hat{y}, \hat{u}, \hat{v})$ will be fed into the light field network. Thus, the network will produce interpolation kernels aligned with $\hat{u}$ and $\hat{v}$.
However for perfect interpolation, we would like the output of the light field network to only depend on $(\hat{s}, \hat{t})$, or for the interpolation kernels to be aligned with $(\hat{s}, \hat{t})$. It can be observed in equations (3) and (4) that the greater the distance $(z_{st} - z_{uv})$, the larger the difference between $(\hat{u}, \hat{v})$ and $(\hat{s}, \hat{t})$, and the \emph{less aligned} the interpolation kernels become with $(\hat{s}, \hat{t})$. 

On the other hand, by learning a re-parameterization of the light field, such that $\secondplane$ is moved towards $\pi^{st}$ (i.e. reducing the distance $(z_{st} - z_{uv})$), we align the  color network's interpolation kernels with $(\hat{s}, \hat{t})$. As in the feature-embedding approach, the finite capacity of the light field MLP will drive the embedding network to learn to map rays intersecting the same point on the textured square to the same point in the latent space---and thus will drive learning of an optimal re-parameterization that leads to good interpolation.

\subsection{Empirical Validation}

\begin{figure}[t]
\centering
\begin{overpic}[width=0.325\linewidth]{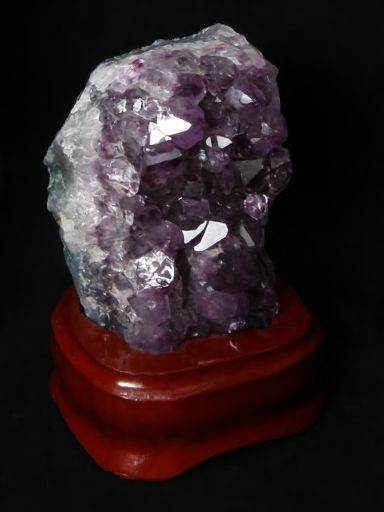}%
\put(0, 50){\textcolor{red}{\rule{0.325\linewidth}{1pt}}}%
\end{overpic}%
\hfill%
\begin{overpic}[width=0.325\linewidth]{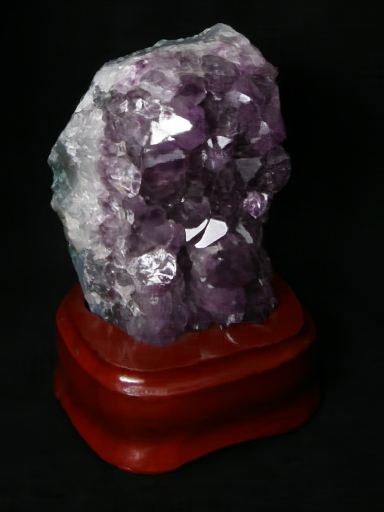}%
\put(0, 50){\textcolor{red}{\rule{0.325\linewidth}{1pt}}}%
\end{overpic}%
\hfill%
\begin{overpic}[width=0.325\linewidth]{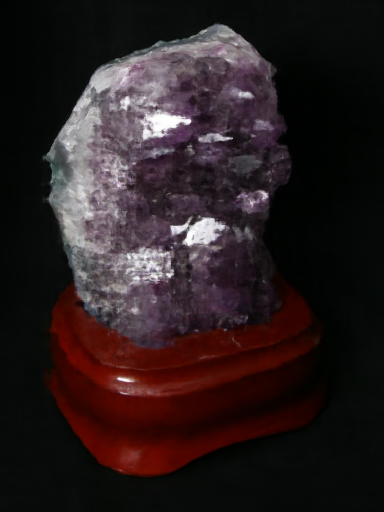}%
\put(0, 50){\textcolor{red}{\rule{0.325\linewidth}{1pt}}}%
\end{overpic}%
\\[0.5mm]%
\includegraphics[width=0.325\linewidth]{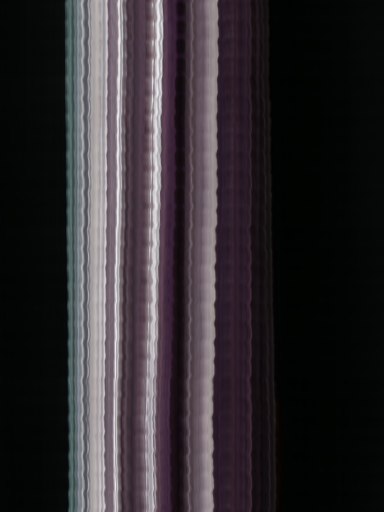}\hfill%
\includegraphics[width=0.325\linewidth]{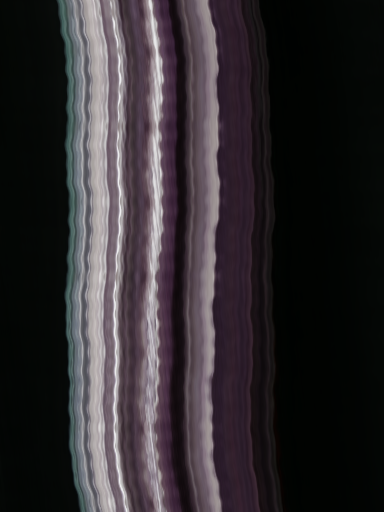}\hfill%
\includegraphics[width=0.325\linewidth]{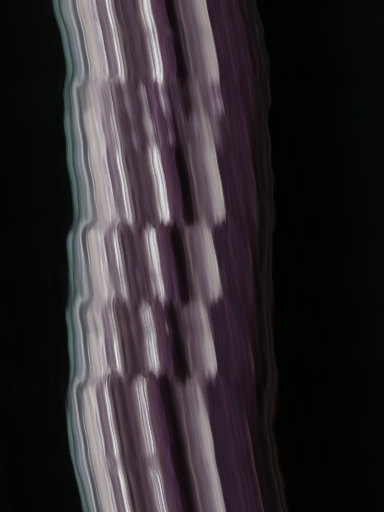}\\[-0.5mm]%
\parbox{0.325\linewidth}{\centering\small $z = 0$}\hfill%
\parbox{0.325\linewidth}{\centering\small $z = 1$}\hfill%
\parbox{0.325\linewidth}{\centering\small $z = 3$}%
\vspace{-1mm}
\caption{
\tb{Effect of initial parameterization.} The \emph{top} row shows predicted images, \emph{bottom} predicted EPIs. Reconstruction becomes progressively worse for a worse initial parameterization ($\secondplane$ moving further and further from the object plane at $z = 0$).
}
\label{fig:reparam_val}
\vspace{-2mm}
\end{figure}

\noindent
We claim that quality of a \emph{baseline} neural light field is correlated with $(z_{st} - z_{uv})$. In order to support this claim, we perform a set of simple experiments with parameterization. In particular, we choose the \textit{Amethyst} scene from the Stanford Light Field dataset \cite{wilburn2005high}, which has very little depth variation. As described in Section~\ref{sec:stanford_data}, each Stanford light field has the object plane $\pi^{st}$ at $z = 0$. We re-parameterize the input light field for $\secondplane$ at $z = 0, 1, 3$, train our model \emph{without} an embedding network, and report validation metrics, as well as showing reconstructed epipolar images. See Figure~\ref{fig:reparam_val} and Table~\ref{tab:reparam_val}.

As hypothesized, the models trained with the object plane closer to $z = 0$ perform better qualitatively and quantitatively. While this is, perhaps, an obvious result, we believe that it is an important one. In particular, it means that light fields with worse initial parameterization are more difficult to learn, and supports the result that learning re-parameterization via local affine transforms vastly improves neural light field interpolation quality.

\subsection{Non-Axis-Aligned Positional Encoding}
\noindent
Positional encoding with non-axis-aligned interpolation kernels (e.g. Gaussian positional encoding \cite{tancik2020fourier}) could be seen as a potential way around the issues discussed above. However, it is important to note that positional encoding with \emph{any fixed set} of interpolation kernels will break down for certain scenes with different depths/depth ranges. For example, this is the case when the interpolation kernels do not align with the light field's color level sets, or there are not enough interpolation kernels to represent high frequencies for particular directions in the light field. In other words, no single-set of interpolation kernels works for all scenes, and ray-space embedding effectively tunes the interpolation per-scene/per-region in ray space,

\begin{table}[t]
\caption{\textbf{Empirical validation of parameterization.} We calculate PSNR, SSIM, and LPIPS for our model without embedding trained on different initial parameterizations. }
\label{tab:reparam_val}
\vspace{-5mm}
\renewcommand{\arraystretch}{1}
\begin{center}
\resizebox{.7\linewidth}{!}{%
\begin{tabular}{@{}l@{\hspace{4\tabcolsep}}c@{\hspace{3\tabcolsep}}c@{\hspace{3\tabcolsep}}c@{}}
  \toprule
  $\secondplane$ location & PSNR$\uparrow$ & SSIM$\uparrow$ & LPIPS$\downarrow$ \\
  \midrule
  $z = 0$ & \textbf{29.631} & \textbf{0.937} & \textbf{0.059} \\
  $z = 1$ & \underline{24.411} & \underline{0.848} & \underline{0.096} \\
  $z = 3$ & 21.491 & 0.790 & 0.146 \\
  \bottomrule
\end{tabular}%
}
\end{center}
\vspace{-2mm}
\end{table}

\section{Embedding Visualization}

\begin{figure}[t]
\centering
\begin{overpic}[width=0.325\linewidth]{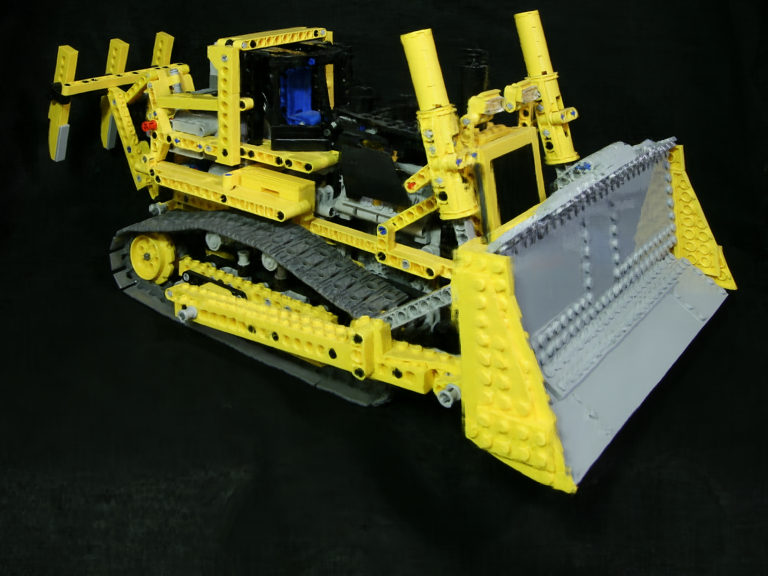}%
\put(0, 38){\textcolor{red}{\rule{0.325\linewidth}{1pt}}}%
\end{overpic}%
\hfill%
\begin{overpic}[width=0.325\linewidth]{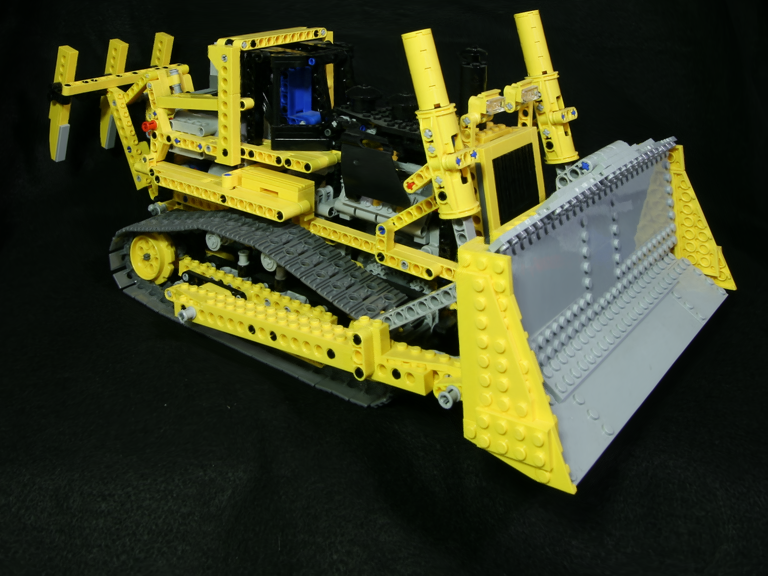}%
\put(0, 38){\textcolor{red}{\rule{0.325\linewidth}{1pt}}}%
\end{overpic}%
\hfill%
\begin{overpic}[width=0.325\linewidth]{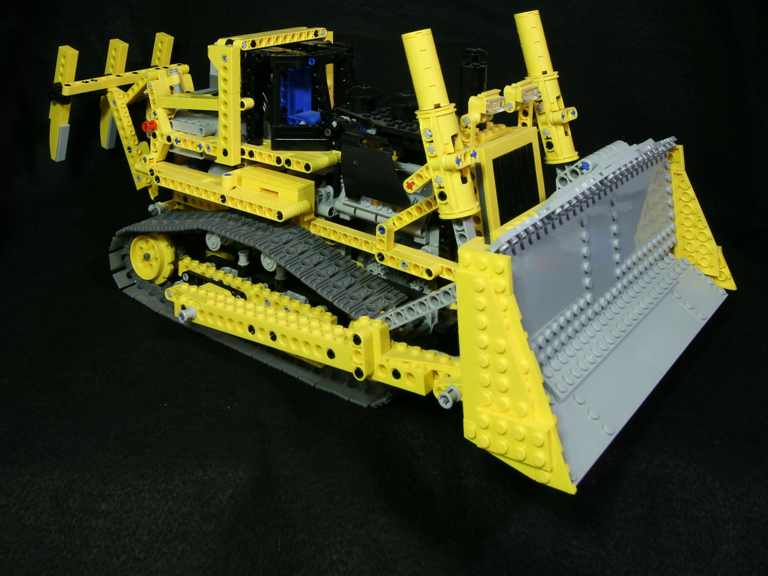}%
\put(0, 38){\textcolor{red}{\rule{0.325\linewidth}{1pt}}}%
\end{overpic}%
\\[0.5mm]%
\includegraphics[width=0.325\linewidth]{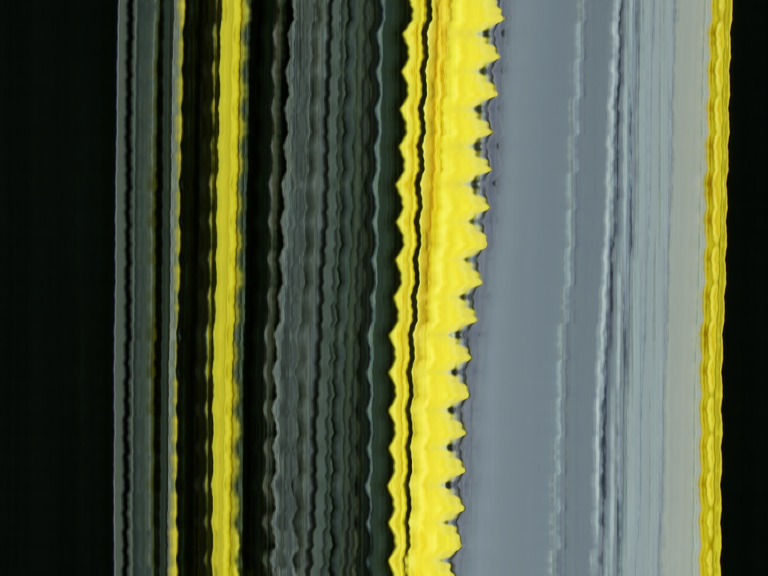}\hfill%
\includegraphics[width=0.325\linewidth]{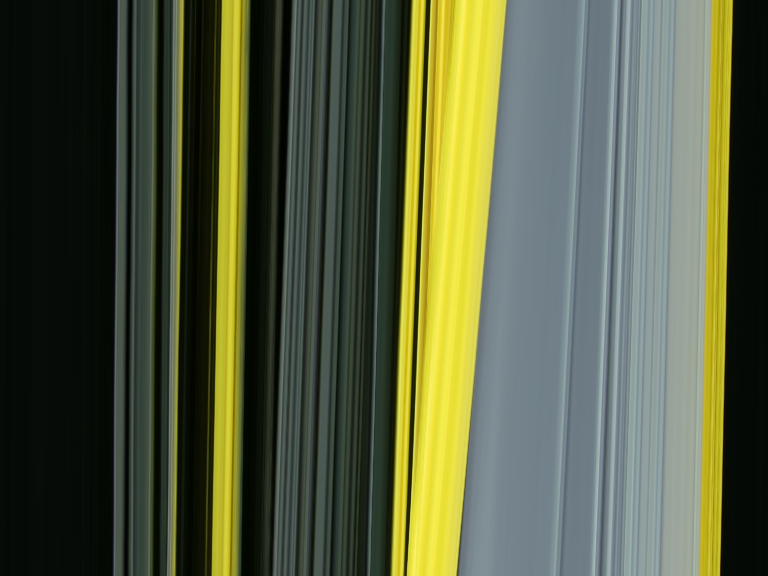}\hfill%
\includegraphics[width=0.325\linewidth]{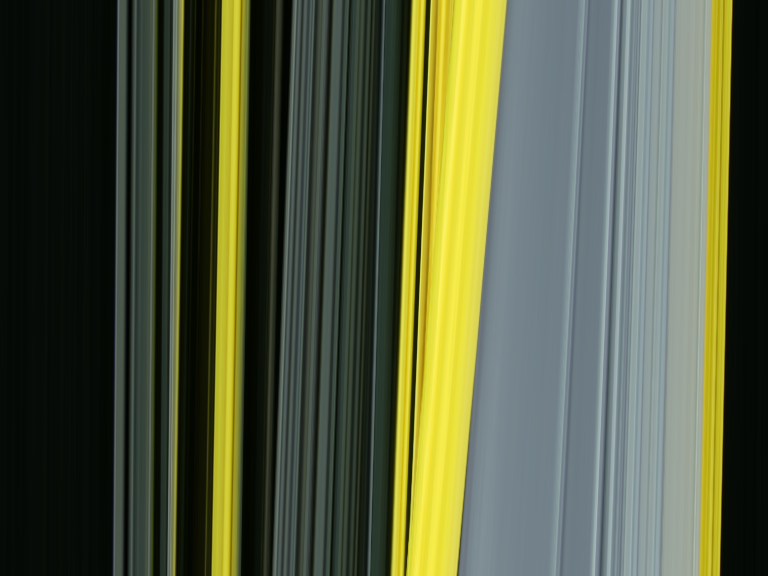}\\[0.5mm]%
\includegraphics[width=0.325\linewidth]{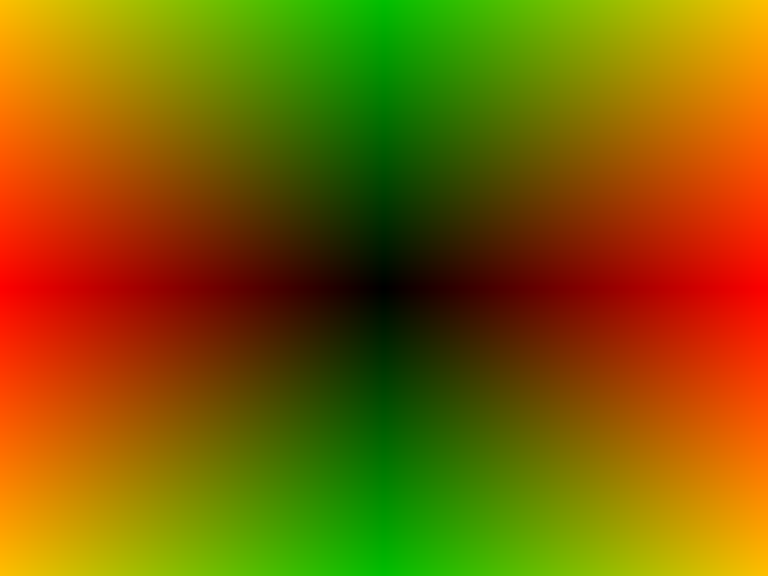}\hfill%
\includegraphics[width=0.325\linewidth]{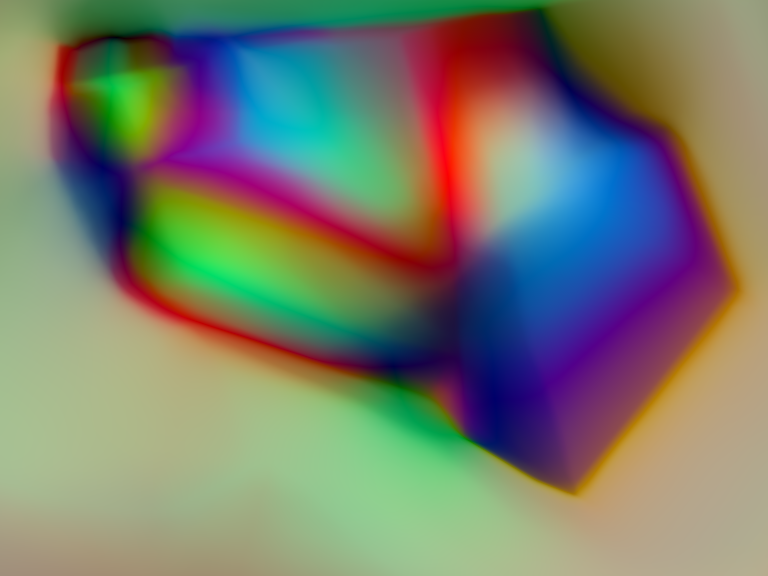}\hfill%
\includegraphics[width=0.325\linewidth]{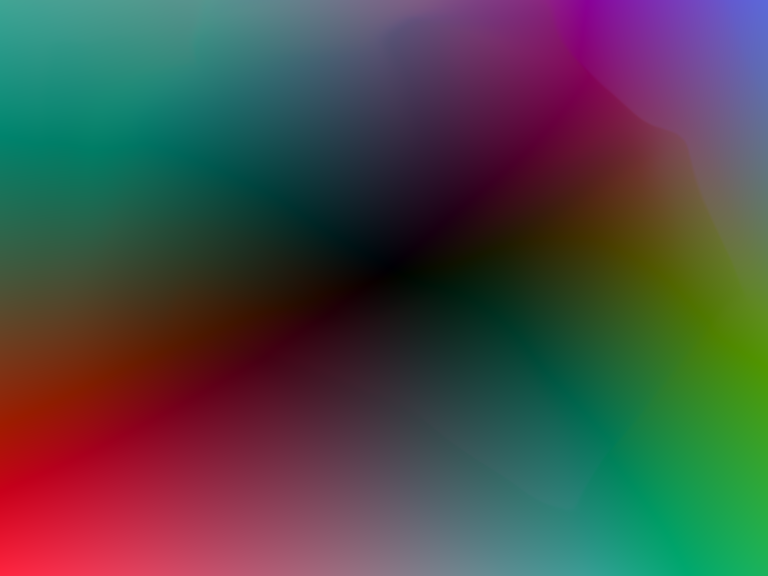}\\[-1mm]%
\parbox{0.325\linewidth}{\centering\scriptsize No Embedding}\hfill%
\parbox{0.325\linewidth}{\centering\scriptsize Feature Embedding}\hfill%
\parbox{0.325\linewidth}{\centering\scriptsize Local Affine Embedding}%
\vspace{-1mm}
\caption{
\tb{Effect of different embeddings}. The \emph{top} row shows predicted images, \emph{middle} predicted EPIs, and \emph{bottom} a visualization of the embedding space for each method.  
}
\label{fig:embedding_results}
\vspace{-4mm}
\end{figure}

\noindent In Figure \ref{fig:embedding_results} we visualize predicted views, predicted EPIs, and the embedded ray-space given as input to the color network for:

\begin{enumerate}
    \item The baseline approach
    \item Feature space embedding
    \item Local affine transform embedding
\end{enumerate}

\noindent For the embedding visualization, RGB colors denote the first three principal components of embedded ray-space.

Note the wiggling artifacts in the EPI predicted without embedding (\emph{left}) . As discussed above, the baseline approach produces axis-aligned interpolation kernels which are ill-suited for interpolating slanted color level sets in the EPI.

The feature embedding network (\emph{center}) learns what is essentially a set of texture coordinates for the object. It registers disparate rays that hit the same 3D points, and interpolates views fairly well, but does not guarantee multi-view consistency as the embedding network is still under-constrained for unobserved views.

Although the transforms themselves are not visualized (it is ray-space \emph{after} applying the transforms that is visualized), the local affine transform network (\emph{right}) predicts a set of transforms that are \emph{largely constant}. This is because the depth range of the scene is limited, and a small set of re-parameterizations works well for all of ray-space. As the network learns a simpler output signal, it naturally interpolates more effectively, even to unobserved views. While the results for feature embedding and local affine transform embedding look similar in Figure \ref{fig:embedding_results}, we encourage readers to visit our web-page. It is far easier to see artifacts of the feature embedding approach in video form.

\begin{figure}[t]
\footnotesize
\centering%
	\mpage{0.2}{%
		\begin{overpic}[width=\linewidth]{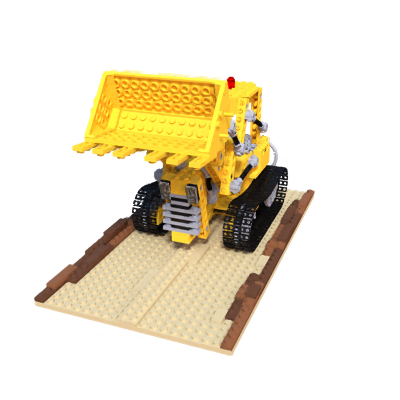}%
			\put(20, 70){\color{red}$\Box$}%
		\end{overpic}\\[-0.5mm]%
		\emph{Lego (Ours)}
	}%
	\hspace{-1mm}\hfill%
	\mpage{0.79}{%
		\mpage{0.3}{%
			\includegraphics[width=\linewidth]{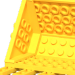}\\[-0.5mm]%
		}\hfill%
		\mpage{0.3}{%
			\includegraphics[width=\linewidth]{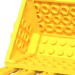}\\[-0.5mm]%
		}\hfill%
		\mpage{0.3}{%
			\includegraphics[width=\linewidth]{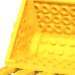}\\[-0.5mm]%
		}
	}\\[1mm]
	\mpage{0.20}{%
	}%
	\hspace{-1mm}\hfill%
	\mpage{0.79}{%
		\mpage{0.24}{%
			GT%
		}\hspace{-1mm}\hfill%
		\mpage{0.24}{%
			Ours%
		}\hspace{-1mm}\hfill%
		\mpage{0.24}{%
			NeRF
		}

	}\\[-2mm]%
\caption{%
We achieve 29.14\,dB on the \emph{Lego} sequence from the NeRF Synthetic dataset \cite{mildenhall2020nerf} with a $32^3$ voxel grid at 800$\times$800 pixel resolution, compared to 27.26\,dB for AutoInt \cite{lindell2021autoint} with 32 sections, and 32.54\,dB for NeRF.
}%
\label{fig:360}
\vspace{-1mm}
\end{figure}

\pagebreak

\section{More Experimental Results}

\noindent
We show per-scene metrics in Tables~\ref{tab:per_scene_realff},~\ref{tab:per_scene_realff_undistort},~\ref{tab:per_scene_shiny}, and ~\ref{tab:per_scene_stanford}. Additionally, we highly encourage readers to visit our project webpage, which contains image comparisons for every scene and video comparisons for a select few scenes.

\topic{360 $^{\circ}$ Scenes}
In Figure~\ref{fig:360} we show preliminary results for our $32^3$ subdivided model with Pl\"ucker parameterization applied to the \emph{Lego} scene in the NeRF Synthetic \cite{mildenhall2020nerf} dataset. We use the same evaluation protocol as in NeRF for this scene. Our PSNR is slightly worse than NeRF, but better than AutoInt for the same grid resolution. Additionally, in some regions, we are able to better recover fine-grained texture on the \emph{Lego} model.

\topic{Student-Teacher Training}
We additionally provide results in Tables~\ref{tab:per_scene_realff} for our method when the input data is augmented with a $10x10$ grid of renderings from a fully trained NeRF. We label this method as ``Ours (w/t),'' or our method with ``student-teacher'' training. With this approach, our method outperforms NeRF quantitatively in terms of PSNR, but at the cost of increased training time. 

\pagebreak

\begin{table*}[t]
\centering
\caption{Per-scene breakdown results from NeRF's Real Forward-Facing dataset~\cite{mildenhall2020nerf}.}
\vspace{-1mm}
\small
\resizebox{\linewidth}{!}{%
\begin{tabular}{%
  @{}%
  l@{\hspace{3\tabcolsep}}%
  c@{\hspace{0.2\tabcolsep}}c@{\hspace{0.9\tabcolsep}}c@{\hspace{0.8\tabcolsep}}c%
  c@{\hspace{0.2\tabcolsep}}c@{\hspace{0.9\tabcolsep}}c@{\hspace{0.8\tabcolsep}}c%
  c@{\hspace{0.2\tabcolsep}}c@{\hspace{0.9\tabcolsep}}c@{\hspace{0.8\tabcolsep}}c@{}%
}
\toprule
\multirow{2}{*}[-0.5ex]{Scene}
& \multicolumn{4}{c}{PSNR$\uparrow$} & \multicolumn{4}{c}{SSIM$\uparrow$} & \multicolumn{4}{c}{LPIPS$\downarrow$} \\
\cmidrule(r){2-5} \cmidrule(lr){6-9} \cmidrule(l){10-13}
& NeRF~\cite{mildenhall2020nerf} & AutoInt~\cite{lindell2021autoint} & Ours & Ours (w/ t) \hspace{-1mm}
& NeRF~\cite{mildenhall2020nerf} & AutoInt~\cite{lindell2021autoint} & Ours & Ours (w/ t) \hspace{-1mm}
& NeRF~\cite{mildenhall2020nerf} & AutoInt~\cite{lindell2021autoint} & Ours & Ours (w/ t) \hspace{-1mm} \\
\midrule
Fern     & \textbf{26.92} & 23.51 & 24.25 & \underline{26.06} & \textbf{0.903} & 0.810 & 0.850 & \underline{0.893} & \textbf{0.085} & 0.277 & 0.114 & \underline{0.104} \\
Flower   & 28.57 & 28.11 & \underline{28.71} & \textbf{28.90} & 0.931 & 0.917 & \textbf{0.934} & \textbf{0.934}    & 0.057 & 0.075 & \textbf{0.038} & \underline{0.053} \\
Fortress & \textbf{32.94} & 28.95 & 31.46 & \underline{32.60} & \textbf{0.962} & 0.910 & 0.954 & \underline{0.961} & \textbf{0.024} & 0.107 & \underline{0.027} & 0.028 \\
Horns    & 29.26 & 27.64 & \textbf{30.12} & \underline{29.76} & 0.947 & 0.908 & \textbf{0.955} & \underline{0.952} & \underline{0.058} & 0.177 & \textbf{0.044} & 0.062 \\
Leaves   & \textbf{22.50} & 20.84 & 21.82 & \underline{22.27} & \underline{0.851} & 0.795 & 0.847 & \textbf{0.855} & \underline{0.103} & 0.156 & \textbf{0.086} & 0.104 \\
Orchids  & \textbf{21.37} & 17.30 & 20.29 & \underline{21.10} & \textbf{0.800} & 0.583 & 0.766 & \underline{0.794} & \underline{0.108} & 0.302 & \textbf{0.103} & 0.113 \\
Room     & \underline{33.60} & 30.72 & 33.57 & \textbf{34.04} & \underline{0.980} & 0.966 & 0.979 & \textbf{0.981} & 0.038 & 0.075 & \textbf{0.037} & \textbf{0.037}    \\
T-rex    & 28.26 & 27.18 & \textbf{29.41} & \underline{28.80} & 0.953 & 0.931 & \textbf{0.959} & \textbf{0.959}    & 0.049 & 0.080 & \textbf{0.034} & \underline{0.040} \\
\bottomrule
\end{tabular}
}
\label{tab:per_scene_realff}
\end{table*}

\begin{table*}[t]
\centering
\caption{Per-scene breakdown results from the \textit{Undistorted} Real Forward-Facing dataset used in NeX~\cite{wizadwongsa2021nex}}
\vspace{-1mm}
\small
\resizebox{0.45\linewidth}{!}{%
\begin{tabular}{%
  @{}%
  l@{\hspace{3\tabcolsep}}%
  c@{\hspace{0.2\tabcolsep}}c@{\hspace{0.9\tabcolsep}}c@{\hspace{0.8\tabcolsep}}c%
  c@{\hspace{0.2\tabcolsep}}c@{\hspace{0.9\tabcolsep}}c@{\hspace{0.8\tabcolsep}}c%
  c@{\hspace{0.2\tabcolsep}}c@{\hspace{0.9\tabcolsep}}c@{\hspace{0.8\tabcolsep}}c@{}%
}
\toprule
\multirow{2}{*}[-0.5ex]{Scene}
& \multicolumn{2}{c}{PSNR$\uparrow$} & \multicolumn{2}{c}{SSIM$\uparrow$} & \multicolumn{2}{c}{LPIPS$\downarrow$} \\
\cmidrule(r){2-3} \cmidrule(lr){4-5} \cmidrule(l){6-7}
& NeX~\cite{wizadwongsa2021nex} & Ours \hspace{-1mm}
& NeX~\cite{wizadwongsa2021nex} & Ours \hspace{-1mm}
& NeX~\cite{wizadwongsa2021nex} & Ours \hspace{-1mm} \\
\midrule
Fern     & \textbf{26.46} & 24.49 & \textbf{0.913} & 0.856 & \textbf{0.068} & 0.107 \\
Flower     & \textbf{29.39} & 28.93 & \textbf{0.947} & 0.937 & 0.041 & \textbf{0.033} \\
Fortress     & \textbf{32.31} & 31.32 & \textbf{0.963} & 0.955 & \textbf{0.024} & 0.026 \\
Horns     & 29.81 & \textbf{29.88} & \textbf{0.959} & 0.952 & \textbf{0.039} & 0.050 \\
Leaves     & \textbf{22.66} & 21.62 & \textbf{0.879} & 0.845 & \textbf{0.082} & 0.082 \\
Orchids     & \textbf{20.51} & 19.93 & \textbf{0.792} & 0.754 & \textbf{0.096} & 0.109 \\
Room     & \textbf{33.40} & 33.24 & \textbf{0.979} & 0.978 & \textbf{0.033} & 0.036 \\
T-rex     & 29.36 & \textbf{29.44} & \textbf{0.965} & 0.963 & 0.037 & \textbf{0.032} \\
\bottomrule
\end{tabular}
}
\label{tab:per_scene_realff_undistort}
\end{table*}

\begin{table*}[t]
\centering
\caption{Per-scene breakdown results from NeX's Shiny Dataset~\cite{wizadwongsa2021nex}}
\vspace{-1mm}
\small
\resizebox{0.45\linewidth}{!}{%
\begin{tabular}{%
  @{}%
  l@{\hspace{3\tabcolsep}}%
  c@{\hspace{0.2\tabcolsep}}c@{\hspace{0.9\tabcolsep}}c@{\hspace{0.8\tabcolsep}}c%
  c@{\hspace{0.2\tabcolsep}}c@{\hspace{0.9\tabcolsep}}c@{\hspace{0.8\tabcolsep}}c%
  c@{\hspace{0.2\tabcolsep}}c@{\hspace{0.9\tabcolsep}}c@{\hspace{0.8\tabcolsep}}c@{}%
}
\toprule
\multirow{2}{*}[-0.5ex]{Scene}
& \multicolumn{2}{c}{PSNR$\uparrow$} & \multicolumn{2}{c}{SSIM$\uparrow$} & \multicolumn{2}{c}{LPIPS$\downarrow$} \\
\cmidrule(r){2-3} \cmidrule(lr){4-5} \cmidrule(l){6-7}
& NeX~\cite{wizadwongsa2021nex} & Ours \hspace{-1mm}
& NeX~\cite{wizadwongsa2021nex} & Ours \hspace{-1mm}
& NeX~\cite{wizadwongsa2021nex} & Ours \hspace{-1mm} \\
\midrule
CD     & 31.92 & \textbf{35.44} & 0.971 & \textbf{0.980} & \textbf{0.028} & \textbf{0.014} \\
Crest     & \textbf{24.78} & 24.48 & \textbf{0.870} & 0.858 & \textbf{0.051} & 0.052 \\
Food     & \textbf{25.61} & 25.21 & \textbf{0.905} & 0.885 & \textbf{0.048} & 0.053 \\
Giants     & \textbf{28.50} & 27.99 & \textbf{0.946} & 0.930 & \textbf{0.038} & 0.039 \\
Lab     & 31.20 & \textbf{34.39} & 0.965 & \textbf{0.982} & 0.031 & \textbf{0.013} \\
Pasta     & \textbf{23.21} & 22.11 & \textbf{0.915} & 0.890 & \textbf{0.045} & 0.065 \\
Seasoning     & \textbf{31.07} & 29.48 & \textbf{0.970} & 0.957 & \textbf{0.028} & 0.045 \\
Tools     & \textbf{29.86} & 28.90 & \textbf{0.974} & 0.968 & \textbf{0.018} & 0.022 \\
\bottomrule
\end{tabular}
}
\label{tab:per_scene_shiny}
\end{table*}

\newcommand{\cmmnt}[1]{}

\begin{table*}[t]
\centering
\caption{Per-scene breakdown results from the Stanford Light Field dataset \cite{wilburn2005high}}
\vspace{-1mm}
\small
\resizebox{\linewidth}{!}{%
\begin{tabular}{%
  @{}%
  l@{\hspace{3\tabcolsep}}%
  c@{\hspace{0.2\tabcolsep}}c@{\hspace{0.9\tabcolsep}}c@{\hspace{0.8\tabcolsep}}c%
  c@{\hspace{0.2\tabcolsep}}c@{\hspace{0.9\tabcolsep}}c@{\hspace{0.8\tabcolsep}}c%
  c@{\hspace{0.2\tabcolsep}}c@{\hspace{0.9\tabcolsep}}c@{\hspace{0.8\tabcolsep}}c@{}%
}
\toprule
\multirow{2}{*}[-0.5ex]{Scene}
& \multicolumn{4}{c}{PSNR$\uparrow$} & \multicolumn{4}{c}{SSIM$\uparrow$} & \multicolumn{4}{c}{LPIPS$\downarrow$} \\
\cmidrule(r){2-5} \cmidrule(lr){6-9} \cmidrule(l){10-13}
& NeRF~\cite{mildenhall2020nerf} & X-Fields~\cite{bemana2020xfields} & NeX~\cite{wizadwongsa2021nex} & Ours
& NeRF~\cite{mildenhall2020nerf} & X-Fields~\cite{bemana2020xfields} & NeX~\cite{wizadwongsa2021nex} & Ours
& NeRF~\cite{mildenhall2020nerf} & X-Fields~\cite{bemana2020xfields} & NeX~\cite{wizadwongsa2021nex} & Ours \\
\midrule
Amethyst      & \underline{39.746} & 37.232 & 39.062 &  \textbf{40.120} & \underline{0.984}\cmmnt{\pz} & 0.982\cmmnt{\pz} & 0.983\cmmnt{\pz} & \textbf{0.985}\cmmnt{\pz}             & 0.026\cmmnt{\pz} & 0.032\cmmnt{\pz} & \underline{0.023}\cmmnt{\pz} & \textbf{0.019}\cmmnt{\pz} \\
Beans         & \textbf{42.519} & 40.911 & 41.776 & \underline{41.659} & \textbf{0.9944} & 0.9931 & \underline{0.9938} & 0.9933                   & \underline{0.014}\cmmnt{\pz} & 0.017\cmmnt{\pz} & 0.016\cmmnt{\pz} & \textbf{0.012}\cmmnt{\pz} \\
Bracelet      & \underline{36.461} & 34.112 & 34.888 & \textbf{36.586} & \underline{0.9909} & 0.9857 & 0.988 & \textbf{0.9913}                   & \underline{0.0094} & 0.0260 & 0.0152 & \textbf{0.0087}       \\
Bulldozer     & \underline{38.968} & 37.350 & 38.131 & \textbf{39.389} & 0.983\cmmnt{\pz} & \underline{0.986}\cmmnt{\pz} & 0.985\cmmnt{\pz} & \textbf{0.987}\cmmnt{\pz}             & 0.063\cmmnt{\pz} & 0.032\cmmnt{\pz} & \underline{0.027}\cmmnt{\pz} & \textbf{0.024}\cmmnt{\pz} \\
Bunny         & \underline{43.370} & 42.251 & 42.722 & \textbf{43.591} & 0.9892 & \underline{0.9894} & 0.9885 & \textbf{0.9900}                   & 0.029\cmmnt{\pz} & \underline{0.022}\cmmnt{\pz} & 0.036\cmmnt{\pz} & \textbf{0.013}\cmmnt{\pz} \\
Chess         & \textbf{41.146} & 37.996 & 39.938 & \underline{40.910} & \underline{0.9915} & 0.9882 & 0.9910 & \textbf{0.9920}                   & 0.028\cmmnt{\pz} & 0.034\cmmnt{\pz} & \underline{0.020}\cmmnt{\pz} & \textbf{0.016}\cmmnt{\pz} \\
Flowers       & \underline{37.910} & 37.590 & 36.982 & \textbf{39.951} & 0.977\cmmnt{\pz} & \underline{0.981}\cmmnt{\pz} & 0.978\cmmnt{\pz} & \textbf{0.984}\cmmnt{\pz}             & 0.076\cmmnt{\pz} & \underline{0.035}\cmmnt{\pz} & 0.036\cmmnt{\pz} & \textbf{0.030}\cmmnt{\pz} \\
Knights       & \textbf{35.978} & 31.491 & \underline{35.678} & 34.591 & \textbf{0.986}\cmmnt{\pz} & 0.974\cmmnt{\pz} & \underline{0.986}\cmmnt{\pz} & 0.982\cmmnt{\pz}             & \textbf{0.0142} & 0.0501 & 0.0168 & \underline{0.0143}       \\
Tarot (Small) & \underline{34.221} & 30.830 & 33.134 & \textbf{36.046} & \underline{0.982}\cmmnt{\pz} & 0.975\cmmnt{\pz} & 0.981\cmmnt{\pz} & \textbf{0.989}\cmmnt{\pz}             & \underline{0.014}\cmmnt{\pz} & 0.033\cmmnt{\pz} & 0.016\cmmnt{\pz} & \textbf{0.006}\cmmnt{\pz} \\
Tarot (Large) & \textbf{24.907} & 24.154 & 22.487 & \underline{24.904} & \underline{0.910}\cmmnt{\pz} & 0.893\cmmnt{\pz} & 0.833\cmmnt{\pz} & \textbf{0.914}\cmmnt{\pz}             & \underline{0.059}\cmmnt{\pz} & 0.074\cmmnt{\pz} & 0.117\cmmnt{\pz} & \textbf{0.039}\cmmnt{\pz} \\
Treasure      & \underline{34.761} & 33.904 & 32.350 & \textbf{37.465} & 0.972\cmmnt{\pz} & \underline{0.977}\cmmnt{\pz} & 0.967\cmmnt{\pz} & \textbf{0.982}\cmmnt{\pz}             & \underline{0.027}\cmmnt{\pz} & 0.041\cmmnt{\pz} & 0.040\cmmnt{\pz} & \textbf{0.019}\cmmnt{\pz} \\
Truck         & \underline{40.723} & 38.883 & 38.292 & \textbf{41.440} & \underline{0.986}\cmmnt{\pz} & 0.984\cmmnt{\pz} & \underline{0.986}\cmmnt{\pz} & \textbf{0.989}\cmmnt{\pz} & 0.087\cmmnt{\pz} & 0.042\cmmnt{\pz}  & \underline{0.038}\cmmnt{\pz} & \textbf{0.033}\cmmnt{\pz} \\
\bottomrule
\end{tabular}
}
\label{tab:per_scene_stanford}
\end{table*}

\end{document}